\documentclass{article}

\usepackage{microtype}
\usepackage{graphicx}
\usepackage{booktabs} 

\usepackage{hyperref}

\newcommand{\cmr}[1]{{\textcolor{black}{#1}}}

\usepackage[accepted]{icml2023}

\usepackage{amsmath}
\usepackage{amssymb}
\usepackage{mathtools}
\usepackage{amsthm}

\usepackage[capitalize,noabbrev]{cleveref}

\theoremstyle{plain}

\theoremstyle{definition}

\theoremstyle{remark}

\usepackage[textsize=tiny]{todonotes}

\usepackage{caption}
\usepackage{subcaption}
\usepackage{multirow}
\usepackage{booktabs}
\usepackage{amssymb}

\renewcommand{\algorithmicrequire}{\textbf{Input:}}  
\renewcommand{\algorithmicensure}{\textbf{Output:}}  

\icmltitlerunning{CLIPood: Generalizing CLIP to Out-of-Distributions}

\begin{document}

\twocolumn[
\icmltitle{CLIPood: Generalizing CLIP to Out-of-Distributions}



\icmlsetsymbol{equal}{*}

\begin{icmlauthorlist}

\icmlauthor{Yang Shu}{equal,tsinghua}
\icmlauthor{Xingzhuo Guo}{equal,tsinghua,iiis}
\icmlauthor{Jialong Wu}{tsinghua}
\icmlauthor{Ximei Wang}{tencent}
\icmlauthor{Jianmin Wang}{tsinghua}
\icmlauthor{Mingsheng Long}{tsinghua}

\end{icmlauthorlist}

\icmlaffiliation{tsinghua}{School of Software, BNRist, Tsinghua University.}
\icmlaffiliation{iiis}{Institute for Interdisciplinary Information Sciences, Tsinghua University.}
\icmlaffiliation{tencent}{Tencent Inc, China. E-mail: Yang Shu \textless{}shu-y18@mails.tsinghua.edu.cn\textgreater{}}

\icmlcorrespondingauthor{Mingsheng Long}{mingsheng@tsinghua.edu.cn}

\icmlkeywords{Machine Learning, ICML}

\vskip 0.3in
]



\printAffiliationsAndNotice{\icmlEqualContribution} 

\begin{abstract}
\cmr{
Out-of-distribution (OOD) generalization, where the model needs to handle distribution shifts from training, is a major challenge of machine learning. Contrastive language-image pre-training (CLIP) models have shown impressive zero-shot ability, but the further adaptation of CLIP on downstream tasks undesirably degrades OOD performances. This paper aims at generalizing CLIP to out-of-distribution test data on downstream tasks. We propose CLIPood, a fine-tuning method that can adapt CLIP models to OOD situations where both domain shifts and open classes may occur on the unseen test data. To exploit the semantic relations between classes from the text modality, CLIPood introduces a new training objective, margin metric softmax (MMS), with class adaptive margins for fine-tuning. To incorporate both pre-trained zero-shot model and fine-tuned task-adaptive model, CLIPood leverages a new optimization strategy, Beta moving average (BMA), to maintain a temporal ensemble weighted by Beta distribution. Experiments on diverse datasets with different OOD scenarios show that CLIPood consistently outperforms existing generalization techniques.
}
\end{abstract}

\section{Introduction}

\begin{figure}[t]
\begin{center}
\centerline{\includegraphics[width=\columnwidth]{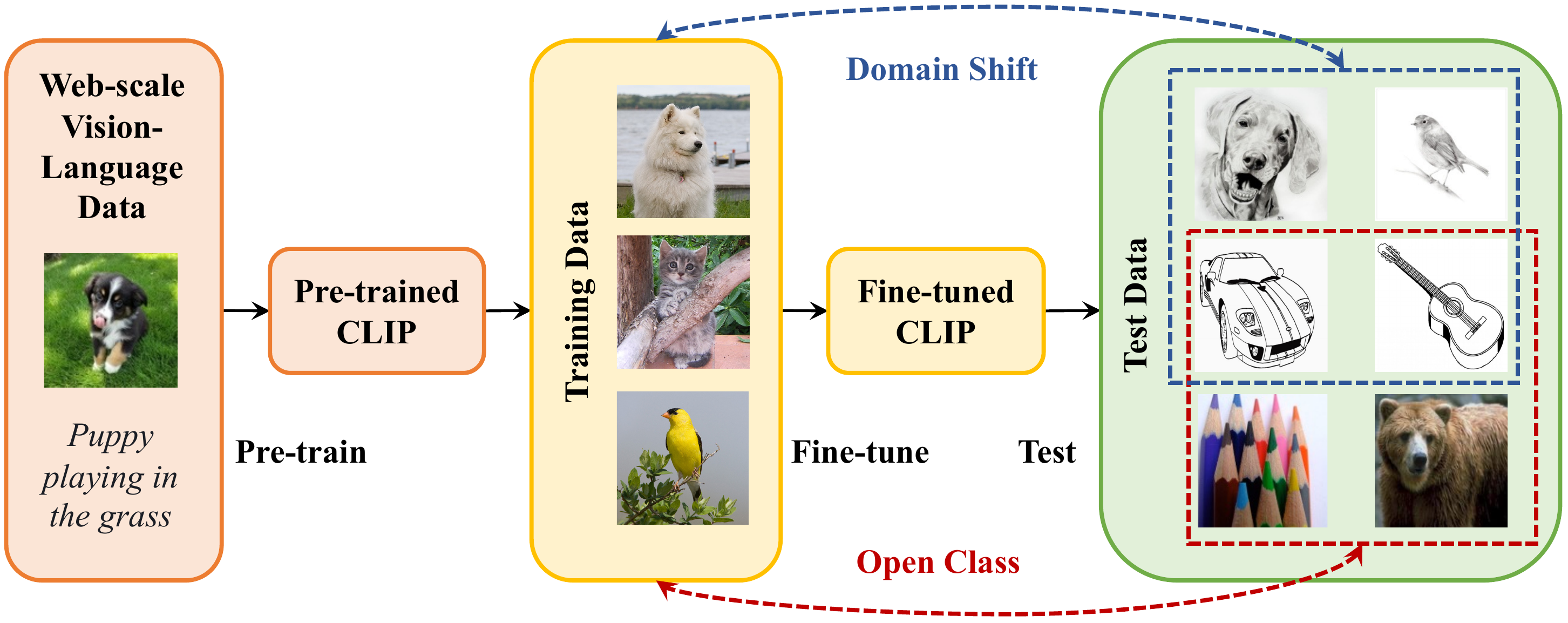}}
\caption{We adapt pre-trained CLIP models on downstream tasks with training data, while maintaining OOD generalization ability to overcome both \emph{domain shift} and \emph{open class}.}
\label{fig:setup}
\vspace{-26pt}
\end{center}
\end{figure}

In a complex and changing open world, machine learning applications inevitably come across the problem of out-of-distribution (OOD) generalization~\cite{cite:ACM2021DeepLearning}, which confronts new tasks with different distributions from the training situation. Even equipped with large-scale pre-trained models and carefully-designed transfer learning algorithms, OOD generalization still remains a significant challenge in the way of developing a reliable machine learning system for the open world~\cite{cite:NIPS2020NaturalDistributionShifts,cite:ICLR2021DomainBed,cite:ICML2021AccuracyOOD}. 
Instead of learning from human-labeled data, recent advances in vision-language pre-training seek to learn from naturally formed supervision of web-scale image-language pairs~\cite{cite:ICML2021CLIP, cite:ICML2021ALIGN}, which enables learning from diverse domains, and recognizing concepts from an open world. As a result, vision-language pre-trained models demonstrate impressive zero-shot learning performance and outperform models trained from only labeled images, which reveals a promising approach toward OOD generalization.

Despite the good zero-shot performance, vision-language models such as Contrastive Language-Image Pretraining (CLIP)~\cite{cite:ICML2021CLIP} achieves OOD generalization in a task-agnostic way. In order for more satisfactory performance on downstream tasks of interest, the pre-trained models still need to utilize task-specific data to make adaptations such as fine-tuning~\cite{cite:ECCV14AnalyzingNN, cite:CVPR14RichFeature}. Although fine-tuning 
achieves better performance than using fixed representations~\cite{cite:CVPR2019DoBetterImageNet}, for CLIP it comes at the cost of OOD generalization: the performance of fine-tuned models may be even worse than zero-shot models on related tasks with distribution shifts~\cite{cite:ICML2021CLIP,cite:Arxiv2021CombinedScaling,cite:CVPR2022Wise}. These results leave OOD generalization an important yet unsolved problem for adapting CLIP models.

Motivated by the promising zero-shot performance, the stronger transfer learning performance against image-only models, and the under-explored challenge of generalization degradation during adaptation, in this paper, we explore the problem of generalizing CLIP models to out-of-distribution data in downstream tasks. \cmr{As shown in Figure \ref{fig:setup}, a more general and challenging setting is proposed, where both types of OOD situations, i.e., \textit{domain shift} (where the training and test data come from different domains), and \textit{open class} (where the test data contain new classes not appearing during training), may occur.} We manage to solve this problem from the view of fine-tuning and seek to handle the dilemma during the adaptation of CLIP models. On the one hand, the pre-trained model should be given the flexibility to fine-tune with the downstream data thus mitigating the gap between upstream and downstream task distributions. On the other hand, since the downstream data are limited and the concrete relationship between the specific training task and the OOD task is unconstrained, the generalization property from large-scale vision-language pre-training should be exploited or maintained to enable safe model adaptation and finally boosts OOD generalization.

Based on this insight, we propose CLIPood, a simple and effective fine-tuning method to improve the OOD generalization ability of CLIP models on downstream tasks. Instead of training an additional classifier for the downstream task, we choose to conduct classification by comparing image embeddings with text embeddings generated from task prompts, which utilizes the knowledge from text modality and keeps the ability of open-class image-text alignment. From the point of the training objective, we propose Margin Metric Softmax (MMS). \cmr{MMS adds an adaptive margin term for each negative class in the metric softmax loss, which is based on its distance from the positive class in the pre-trained text space. By adding such a margin}, MMS explores semantic relationships from vision-language pre-training to boost the OOD generalization during fine-tuning. From the point of model optimization, we propose Beta Moving Average (BMA). Tailored to the fine-tuning trajectory of CLIP models, where the pre-trained model has good zero-shot performance and the adapted model has the knowledge of specific downstream tasks, BMA maintains a temporal ensemble \cmr{for the intermediate models in the fine-tuning trajectory, and the contributions of models from different training steps are determined according to their corresponding probability in the Beta distribution.}

The contributions of the paper can be summarized as:
\begin{itemize}
    \vspace{-5pt}
    \setlength{\itemsep}{0.2em}
    \item We aim at an \cmr{under-explored problem} of generalizing CLIP models to out-of-distributions, and propose a more general and challenging in-the-wild setting where both domain shift and open class occur on test data. 
    \item We propose CLIPood, a simple and effective fine-tuning method of CLIP. Based on metric softmax fine-tuning, CLIPood proposes a new \cmr{training objective} Margin Metric Softmax, and a new model \cmr{optimization strategy} Beta Moving Average to boost OOD generalization performance on downstream tasks.
    \item We conduct experiments on various datasets with different OOD scenarios, including domain shift, open class and co-occurrence of both. Experimental results show that the proposed CLIPood method \cmr{consistently outperforms} existing generalization techniques.
\end{itemize}

\begin{figure*}[tbp]
  \centering
  \includegraphics[width=0.85\textwidth]{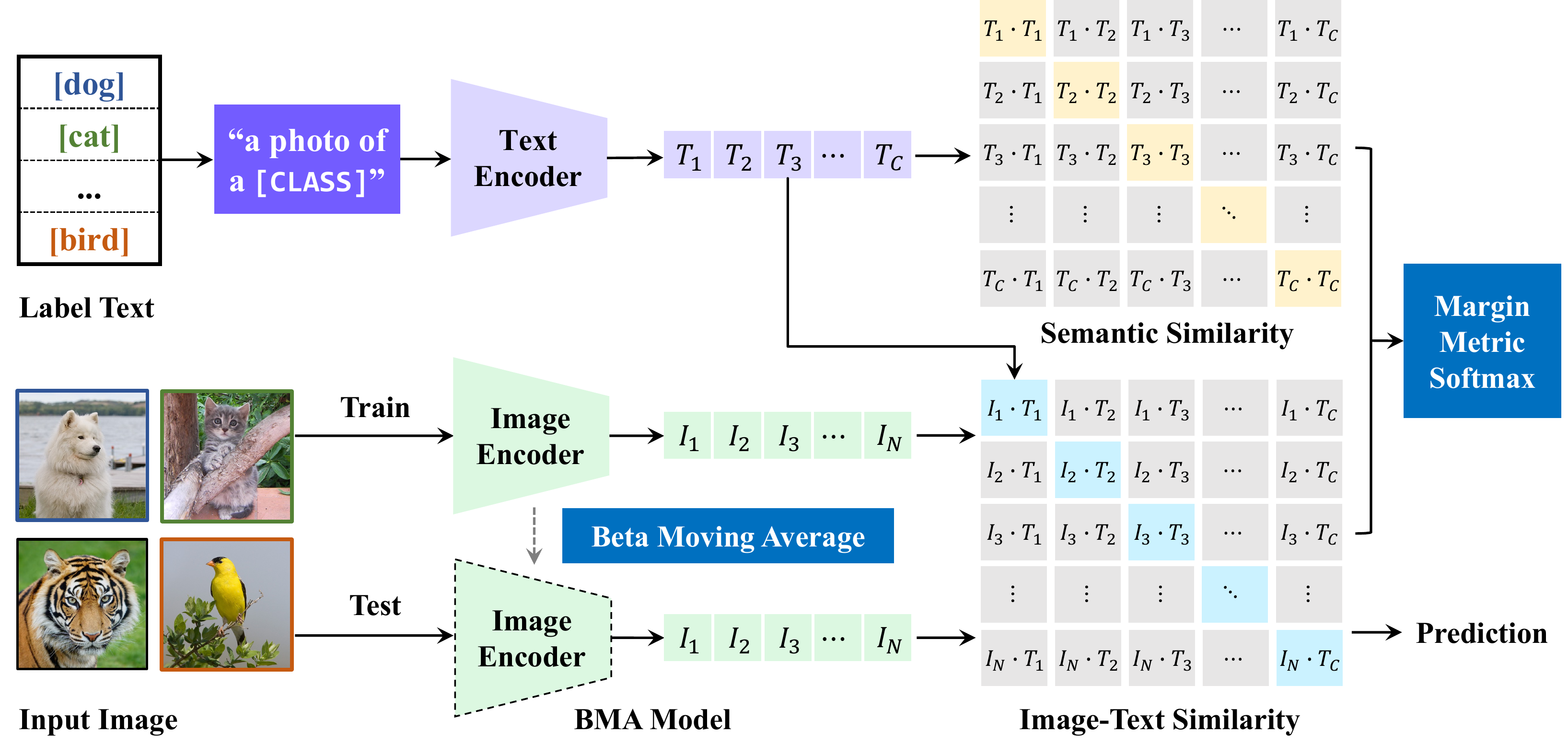}
  \caption{Overview of the proposed CLIPood method. CLIPood compares image embeddings with class text embeddings. Margin Metric Softmax is introduced to exploit semantic relationships between classes. Moreover, a Beta Moving Average model is maintained for prediction, which incorporates both the pre-trained zero-shot model and the fine-tuned model.}
   \label{fig:arch}
   \vspace{-5pt}
\end{figure*}

\section{Related Work}

\textbf{Vision-Language Pre-training.} Vision-language models connect images and texts through a common embedding space to enable cross-modal learning~\cite{cite:NIPS2013Devise,cite:NIPS2013ZeroShotCrossModal,cite:ICCV2013WriteaClassifier}. Recent advances employ architectures with better representation learning abilities such as Transformer~\cite{cite:NIPS2017Transformer} and web-scale training datasets and build stronger vision-language pre-trained models. One type of the  approach learns the common embedding space by masked language modeling or masked region prediction~\cite{cite:NIPS2019VLBert,cite:EMNLP2019LXMERT,cite:ICLR2020VLBERT,cite:ICML2021Vilt}. In this paper, we focus on another typical type of contrastive language-image pre-training such as CLIP~\cite{cite:ICML2021CLIP} and ALIGN~\cite{cite:ICML2021ALIGN}. Recent research also seeks to improve the pre-training paradigm, such as using additional supervision~\cite{cite:Arxiv2021DeCLIP,cite:ECCV2022Slip}, employing pre-trained image encoders~\cite{cite:CVPR2022Lit}, and adding cross-modal and in-modal consistency constraints~\cite{cite:NIPS2022Cyclip}. In this paper, instead of designing better pre-training techniques, we aim at utilizing recent advances in vision-language pre-trained models such as CLIP and achieving better OOD performance.

\textbf{Out-of-Distribution (OOD) Generalization.} Research on OOD generalization aims to improve the performance of the model on new data with different distributions from the training data. One typical research topic domain generalization explores domain distribution shifts, which trains the model with source-domain data and aims at achieving high performance in unseen target domains~\cite{cite:ECCV12undoing,cite:ICML13DICA}. Most domain generalization methods focus on the training strategies on source domains, including cross-domain feature alignment~\cite{cite:ECCV18CIAN}, decomposing domain-specific and domain-common knowledge~\cite{cite:ICML20CSD,cite:ECCV20DMG}, meta-learning over domains~\cite{cite:AAAI18MLDG,cite:NIPS18MetaReg}, designing data-augmentation tasks~\cite{cite:NIPS2018ADA, cite:CVPR19JiGen}, \cmr{and weight ensemble~\cite{cite:NIPS2021SWAD}}. Besides domain distribution shift, recent research also explores open classes in heterogeneous domain generalization~\cite{cite:ICML2019FeatureCritic} and open domain generalization~\cite{cite:CVPR2021ODG}. However, these settings still cannot ensure training-free generalization or can only perform open-class detection, which is limited by the closed-set property of the pre-trained models. \citet{cite:ICLR2021DomainBed} compare various methods fairly on the same benchmark and show that only focusing on algorithm design may not fully address the OOD generalization problem.

Vision-language pre-trained models such as CLIP exhibit impressive zero-shot generalization ability to the open world, which opens a new path towards stronger OOD generalization. Despite the good zero-shot performance, research finds that further adapting CLIP models with task-specific data comes at the cost of OOD generalization ability~\cite{cite:ICML2021CLIP,cite:CVPR2022Wise}. Recent advances explore improving the OOD generalization of CLIP models on the downstream tasks by adapter learning~\cite{cite:Arxiv2021CLIP-Adapter,cite:Arxiv2021TipAdapter}, model ensemble~\cite{cite:CVPR2022Wise}, test-time adaptation~\cite{cite:NIPS2022TestTimePrompt}, and prompt learning~\cite{cite:IJCV2022CoOp,cite:CVPR2022ProDA,cite:CVPR2022CoCoOp}. \cmr{Compared with most of the existing works on prompt learning or adapter learning, we focus on the aspect of fine-tuning CLIP models, which is a simple and common practice for transfer learning but an under-explored point for generalizing CLIP to out-of-distributions. Compared with some recent fine-tuning-based methods~\cite{cite:CVPR2022Wise,cite:Arxiv2022LPFT,cite:Arxiv2023FLYP}, we propose a novel design from the aspects of both training objectives and model optimization.} We consider two types of OOD situations with domain shift and open class and propose to solve a more general and challenging problem where both these two OOD situations may appear in unseen test data.

\section{Method}

\subsection{Generalizing CLIP to Out-of-Distributions}

\textbf{CLIP Models.} In this paper, we focus on generalizing the vision-language pre-trained model CLIP~\cite{cite:ICML2021CLIP} to OOD distributions. Instead of supervision from human labels, CLIP learns directly from the raw texts about images. During the training time, CLIP jointly trains an image encoder $g_I(\cdot)$ and a text encoder $g_T(\cdot)$ in a contrastive learning way to predict the correct pairings of the image-text training samples. At the test time, the names of the target dataset's classes are embedded by the learned text encoder to synthesize a zero-shot classifier for the target task.

\textbf{Problem Setup.} We explore the problem of generalizing CLIP models to out-of-distributions. Given a pre-trained CLIP model, it is first adapted with some training data $\mathcal{S}=\{(\mathbf{x}, y)\}$ of the downstream task, with class label $y \in \mathcal{Y}$. The adapted model should achieve good generalization on related but \textit{out-of-distribution} test data $\mathcal{T} = \{(\mathbf{x}', y')\}$, with the class label $y' \in \mathcal{Y}'$. We explore two OOD scenarios in this paper. In the \textit{domain shift} scenario, we have $P(\mathbf{x}, y) \neq P(\mathbf{x}', y')$, which means the test data may come from domains with different distributions. In the \textit{open class} scenario, we have $\mathcal{Y} \neq \mathcal{Y}'$, which means the test data may contain new classes not appearing in the training data. Instead of the controlled setting that considers two scenarios separately, we propose an even more challenging \emph{in-the-wild} setting where both types of distribution shifts may occur simultaneously in the test data.

\subsection{CLIP Fine-tuning}

In this paper, we aim to design a method that further fine-tunes and adapts CLIP models to downstream tasks. In standard fine-tuning, a linear classifier $\mathbf{W}=\{\mathbf{w}_c\}_{c=1}^{C}$ is employed to fit for the new task, with each learnable parameter vector $\mathbf{w}_c$ representing a new class $c$. The classifier makes predictions on the image feature $g_I\left(\mathbf{x}\right)$ 
, and outputs the probability that the sample $\mathbf{x}$ belongs to the class $y$ as:
\begin{equation}
    P(y|\mathbf{x}) = \frac{\exp{\left( \mathbf{w}_y \cdot g_I\left(\mathbf{x}\right) \right)}}{\sum_{c=1}^{C}{\exp{\left( \mathbf{w}_c \cdot g_I\left(\mathbf{x}\right) \right)}}}.
\end{equation}
Then training signals, e.g. cross-entropy losses, are used to simultaneously train the classifier $\mathbf{W}$ and fine-tune the image encoder $g_I(\cdot)$ on the downstream task.

For CLIP models, applying the standard fine-tuning strategy only transfers the knowledge in the image modality, which discards the knowledge in the text modality and breaks the connection between them. This may decrease the generalization ability benefiting from image-text alignment. Besides, the added classifier is tailored to the training dataset, making it hard to generalize to unseen classes. 
Therefore, we propose to perform a vision-language fine-tuning strategy on CLIP to enhance its OOD generalization ability.

Inspired by the zero-shot prediction protocol of CLIP, for each class $c$, we generate a text prompt $\mathbf{t}_c$ describing it, such as ``a photo of a \texttt{[CLASS]},'' where the \texttt{[CLASS]} token is replaced by the name of class $c$. We then get the text embedding of each class $\mathbf{T}_c=g_T(\mathbf{t}_c)$ extracted by the text encoder. For making predictions, we compare the image embedding $\mathbf{I}_{\mathbf{x}}=g_I(\mathbf{x})$ of input image $\mathbf{x}$ with the class text embeddings. Concretely, the probability that a sample $\mathbf{x}$ belongs to class $y$ is computed as a metric softmax:
\begin{equation} \label{eqn:clip-metric}
    P(y|\mathbf{x}) = \frac{\exp{\big( S \left(\mathbf{I}_\mathbf{x}, \mathbf{T}_{y} \right) / \tau \big)} } { \sum_{c=1}^{C} \exp{\big( S \left(\mathbf{I}_\mathbf{x}, \mathbf{T}_{c} \right) / \tau \big)} },
\end{equation}
where $S(\cdot,\cdot)$ is a similarity metric between embeddings, and $\tau$ is a temperature. We follow the training protocol in CLIP to use the cosine similarity. We can fine-tune the CLIP model with such predictions, which utilizes knowledge in both image and text modalities and preserves the alignment between them. Different from the pre-training stage with abundant and diverse image-text pairs, in downstream scenarios with images and class text prompts, image patterns are still rich, but the diversity of text corpus is limited.
Fine-tuning the text encoder would ruin the semantic relations of concepts pre-trained from a diverse world and overfit to sided knowledge of downstream training classes, which degrades the performance in open class scenarios.
Thus, we fine-tune the image encoder to adapt to downstream tasks and freeze the text encoder to avoid representation collapse. To further exploit the inherent unequal relations of classes, we next introduce our proposed Margin Metric Softmax.

\subsection{Margin Metric Softmax}

When directly applying cross-entropy loss with the prediction in Equation~\eqref{eqn:clip-metric} to update the model, it enforces that the similarity between the image embedding $\mathbf{I}_{\mathbf{x}}$ of each sample $\mathbf{x}$ and the text embedding $\mathbf{T}_{y}$ of the correct class $y$ is higher than those $\mathbf{T}_{c}$ of other false classes $c$. This aligns the image embedding with the correct text embedding but treats all the false equally, which ignores the potential semantic relations between classes. On the other hand, the pre-trained text modality contains more detailed semantic knowledge, which quantifies the semantic relationships between texts in detail other than just discriminating between classes. Thus, we propose to explore such knowledge to enhance the generalization during vision-language fine-tuning. 

\begin{figure}[t]
\begin{center}
\centerline{\includegraphics[width=0.8\columnwidth]{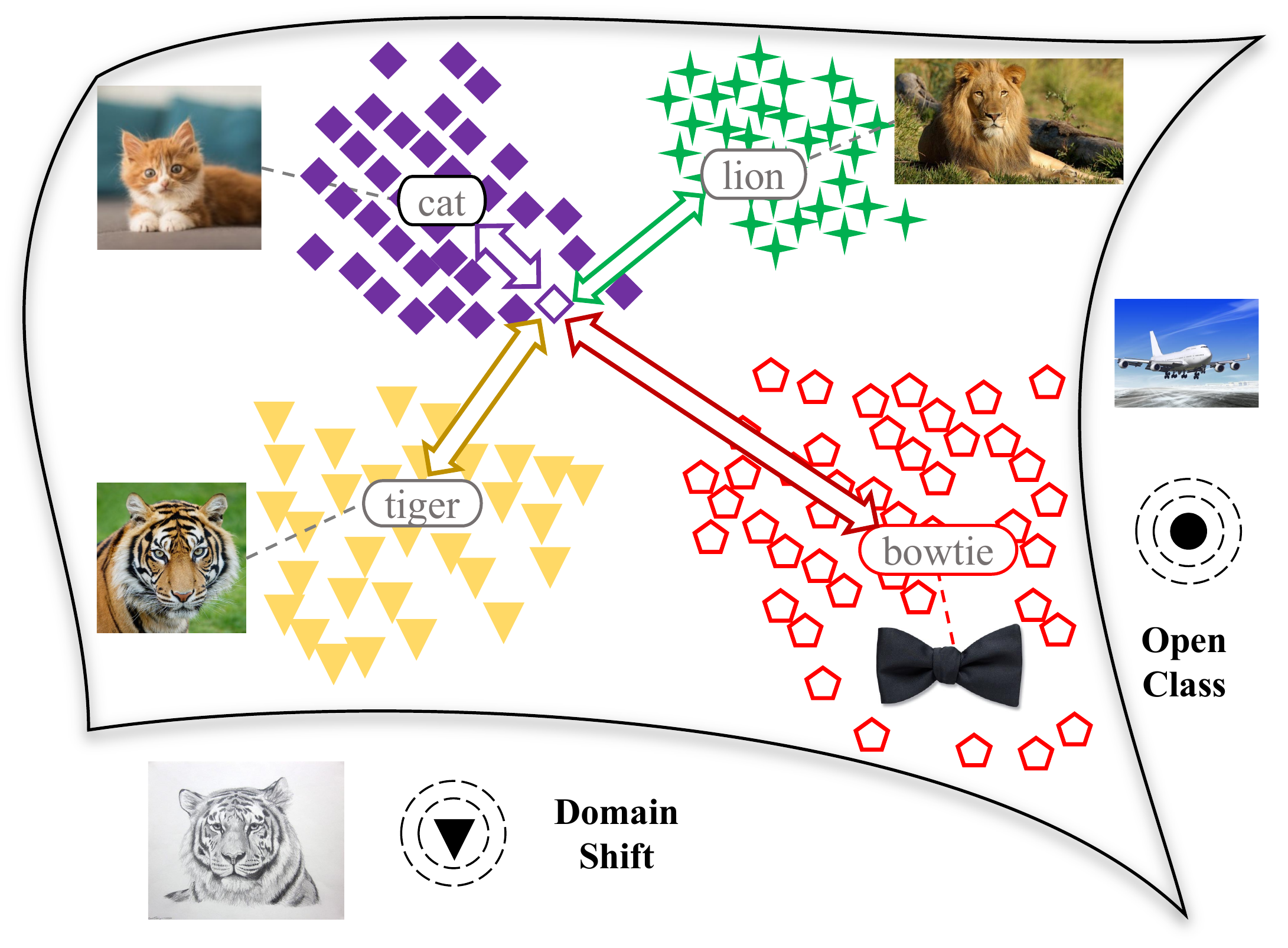}}
\vspace{-5pt}
\caption{The illustration of Margin Metric Softmax (MMS), where the hollow diamond at the center represents the image embedding $\mathbf{I}_\mathbf{x}$. Since $D \left(\mathbf{T}_y, \mathbf{T}_{c} \right)$ varies across classes, the \textit{adaptive margin} is attained, preserving the inherent unequal relations of classes.}
\vspace{-20pt}
\end{center}
\end{figure}

For each training sample $(\mathbf{x},y)$, we newly propose the Margin Metric Softmax (MMS) loss as:
\begin{equation}\label{eqn:adaptive-margin}
    \mathcal{L} = - \log \frac{\exp{\big( S \left(\mathbf{I}_\mathbf{x}, \mathbf{T}_{y} \right) / \tau \big)}}{ \sum_{c=1}^{C} \exp{\big( ( S \left(\mathbf{I}_\mathbf{x}, \mathbf{T}_{c} \right) + \lambda \cdot D \left(\mathbf{T}_y, \mathbf{T}_{c} \right) ) / \tau \big)}}.
\end{equation}
Here, $D \left(\mathbf{T}_y, \mathbf{T}_{c} \right) $ represents the distance between the text embeddings of classes $y$ and $c$, instantiated naturally as: 
\begin{equation}
    D \left(\mathbf{T}_y, \mathbf{T}_{c} \right) = 1 - S \left(\mathbf{T}_y, \mathbf{T}_{c} \right).
\end{equation}
The term $\lambda \cdot D \left(\mathbf{T}_y, \mathbf{T}_{c} \right)$ serves as an \textit{adaptive margin} for each $S \left(\mathbf{I}_\mathbf{x}, \mathbf{T}_{c} \right)$ in the loss. $\lambda$ is a hyper-parameter that trades off the image-text similarity and the class-embedding distance. Note that $D \left(\mathbf{T}_y, \mathbf{T}_{y} \right)=0$, thus these margin terms enforce that the similarity with the correct text label is higher than those with false text labels by an adaptive margin, which strengthens the image-class alignment. Different from a fixed margin for all classes, the adaptive term $D \left(\mathbf{T}_y, \mathbf{T}_{c} \right)$ implies that when the semantic distance between classes $y$ and $c$ is small, the margin term only makes a small difference, but when the semantic distance is larger (indicating that this false text label is much more different from the correct text label in semantic meanings), the margin term pushes the image embedding further way from such a false text label. In this way, MMS exploits the more detailed knowledge of semantic relations in the pre-trained text modality to achieve a better image-text cross-modal alignment and enhance the generalization of the model during vision-language fine-tuning.

\subsection{Beta Moving Average}

\cmr{Despite generally better performance on downstream tasks}, fine-tuning pushes the model far away from the pre-trained one at the risk of catastrophic forgetting and representation collapse. This may hurt OOD generalization ability, especially considering that the CLIP pre-trained model itself is a good zero-shot learner. Additional regularization can be applied to preserve the pre-trained knowledge. However, it always needs task-specific designs and careful tuning of extra hyper-parameters, which hinders its flexibility in real-world applications. In this paper, from the perspective of model optimization, we newly propose Beta Moving Average (BMA) to maintain the benefits of both sides. 

Consider a fine-tuning procedure of $T$ training steps, we can get a trajectory of models $\{\theta_{t}\}_{t=0}^{T}$, where ${\theta}_{0}$ is the pre-trained model and ${\theta}_{t}$ is the model at the $t$-th step. We aim to compute the temporal ensembling $\theta^{\text{TE}}$ of the training procedure as a weighted average of intermediate models:
\begin{equation}
    \theta^{\text{TE}} = \sum_{t=0}^{T} \frac{\alpha_t}{\sum_{k=0}^{T}\alpha_k} \cdot \theta_{t},
\end{equation}
where $\alpha_t$ determines the contribution of each model $\theta_{t}$. During CLIP fine-tuning, the checkpoints near the pre-trained model (with a smaller step $t$) keep more knowledge from large-scale pre-training, which results in better generalization on various domains and classes in a task-agnostic manner but lacks task-specific knowledge. In contrast, the checkpoints near the fine-tuned model (with a larger step $t$) has been injected more task-related knowledge through training, but the generalization of such knowledge is not guaranteed due to the unknown relationship with the OOD test data. Since both sides contribute to the final OOD generalization performance, we want to strengthen the influence of the models near the two sides with a distribution prior on their weights. We propose Beta Temporal Ensemble which normalizes the training steps to $(0,1)$ and samples from a Beta distribution $\text{Beta}(\beta,\beta)$ to determine the weight of each model as its corresponding probability in the distribution:
\begin{equation}\label{eqn:Beta Weight}
    \alpha_t = \text{Beta}(\beta,\beta) \left( \frac{t+0.5}{T+1} \right).
\end{equation}
Here, $\beta$ is a hyper-parameter and we choose $\beta < 1$ to focus more on pre-trained and fine-tuned models.

\begin{figure}[t]
\begin{center}
\centerline{\includegraphics[width=\columnwidth]{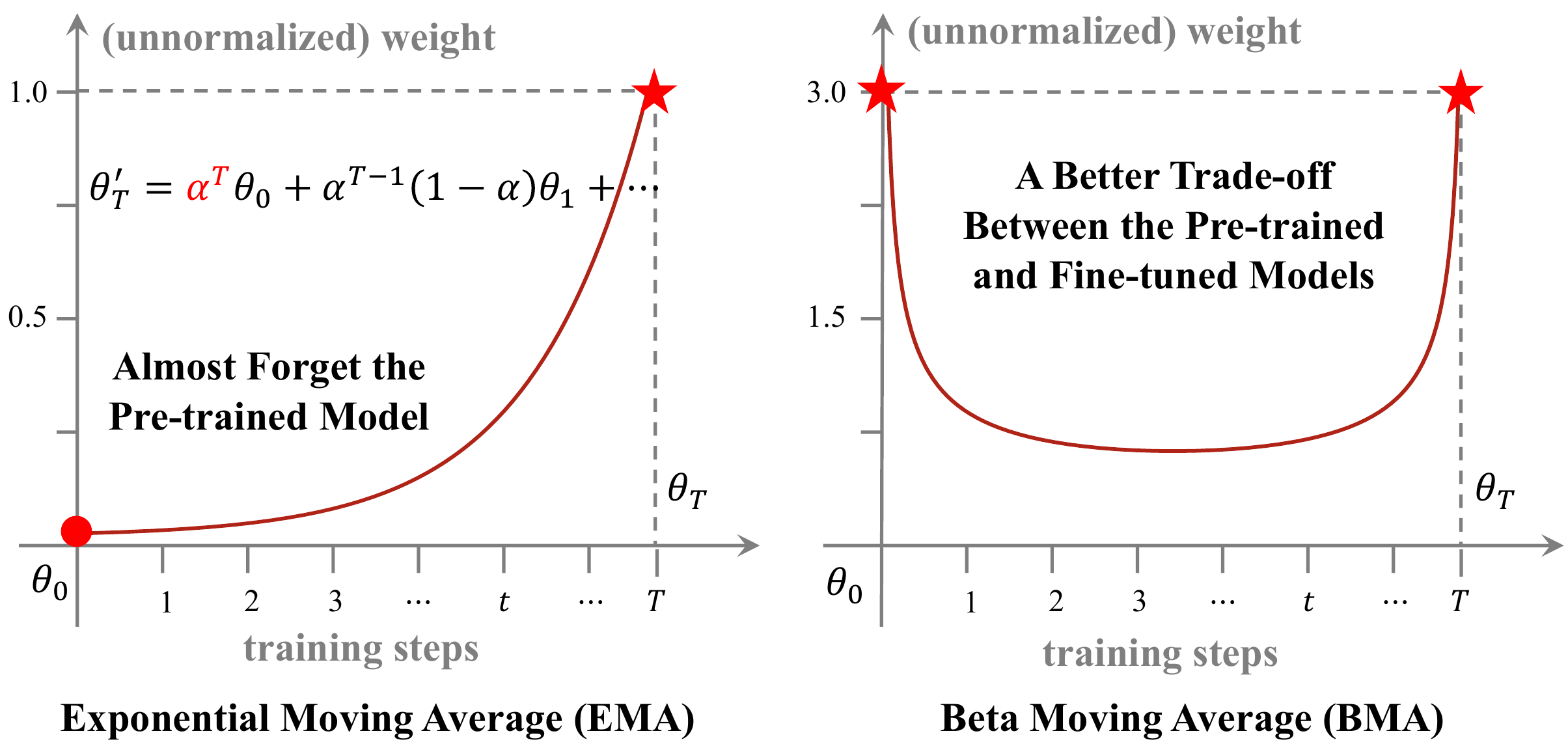}}
\caption{A comparison between Exponential Moving Average (EMA) and Beta Moving Average (BMA), in which the first term of EMA is $\alpha^T\theta_0$ over the $T$ training step and $\theta_0$ is the pre-trained model. Since $\alpha^T \to 0$ when $0 <\alpha < 1$, the fine-tuned model with EMA will almost forget the knowledge of the pre-trained model.}
\vspace{-15pt}
\label{fig:bma}
\end{center}
\end{figure}

Directly performing the temporal model ensemble requires saving many snapshots of the model on the training trajectory, which greatly increases the storage cost. To mitigate this problem, we further adjust Beta Temporal Ensemble into Beta Moving Average, which computes the average of current models on the fly. We maintain a moving average model $\theta^{\text{BMA}}$ and at each time step $t$, the current model $\theta_{t}$ is added into $\theta^{\text{BMA}}_t$ to update the moving average:
\begin{equation}\label{eqn:BMA}
    \theta^{\text{BMA}}_{t} = \frac{\sum_{k=0}^{t-1}\alpha_k}{\sum_{k=0}^{t}\alpha_k} \cdot \theta^{\text{BMA}}_{t-1} + \frac{\alpha_t}{\sum_{k=0}^{t}\alpha_k} \cdot \theta_{t}.
\end{equation}
We present a comparison between common-used Exponential Moving Average (EMA) and the proposed Beta Moving Average (BMA) in Figure \ref{fig:bma}.

The whole architecture of the CLIPood model is shown in Figure~\ref{fig:arch}. In Algorithm~\ref{alg:BMA}, we show the overall training procedure of the proposed CLIPood method. The pre-trained model is fine-tuned with Margin Metric Softmax (MMS), and at each step of the fine-tuning, the Beta Moving Average (BMA) of the models is computed and updated on the fly. The final BMA model is stored to make OOD predictions.

\begin{algorithm}[H]
\caption{Training Procedure of CLIPood}
\begin{algorithmic}
\label{alg:BMA}
\REQUIRE Pre-trained CLIP model $\theta_{0}$, learning rate $\eta$
\STATE Initialize the BMA model \(\theta^{\text{BMA}}_{0}\gets\theta_0\)
\FOR{\(t\in[1,T]\)}
  \STATE Sample data \(\{(\mathbf{x},{y})\}\) from the training set $\mathcal{S}$
  \STATE Calculate MMS loss $\mathcal{L}$ as in Eq.~\eqref{eqn:adaptive-margin}
  \STATE Update model parameters $\theta_{t}\gets\theta_{t-1}-\eta\nabla_{\theta_{t-1}}\mathcal{L}$
  \STATE Calculate $\alpha_t$ of the current model as in Eq.~\eqref{eqn:Beta Weight}
  \STATE Update the BMA model $\theta^{\text{BMA}}_{t}$ as in Eq.~\eqref{eqn:BMA}
\ENDFOR
\ENSURE The final BMA model $\theta^{\text{BMA}}_T$
\end{algorithmic}
\end{algorithm}

\begin{table*}[t]
\caption{Accuracy on the DomainBed benchmark with domain shift.}
\label{table:domainbed}
\vspace{-10pt}
\begin{center}
\begin{small}
\begin{sc}
\begin{tabular}{lccccccc}
\toprule
Method & Backbone & PACS & VLCS & OfficeHome & TerraInc & DomainNet & Avg. \\
\midrule
ERM & ResNet & 85.5 & 77.5 & 66.5 & 46.1 & 40.9 & 63.3 \\
CORAL~(\citeyear{cite:ECCV2016DeepCoral}) & ResNet & 86.2 & 78.8 & 68.7 & 47.6 & 41.5 & 64.6 \\
Zero-shot & CLIP   & 96.2 & 81.7 & 82.0 & 33.4 & 57.5 & 70.2 \\
ERM     & CLIP & 96.1$_{\pm 0.5}$ & 83.0$_{\pm 0.2}$ & 83.3$_{\pm 0.3}$ & \textbf{60.9}$_{\pm 0.2}$ & 59.9$_{\pm 0.1}$ & 76.7$_{\pm 0.2}$ \\
MIRO~(\citeyear{cite:Arxiv2022MIRO})    & CLIP & 95.6 & 82.2 & 82.5 & 54.3 & 54.0 & 73.7 \\
DPL~(\citeyear{cite:Arxiv2022DomainPrompt})     & CLIP & \textbf{97.3} & 84.3 & 84.2 & 52.6 & 56.7 & 75.0 \\
CLIPood & CLIP & \textbf{97.3}$_{\pm 0.1}$ & \textbf{85.0}$_{\pm 0.4}$ & \textbf{87.0}$_{\pm 0.2}$ & 60.4$_{\pm 0.7}$ & \textbf{63.5}$_{\pm 0.1}$ & \textbf{78.6}$_{\pm 0.1}$ \\

\bottomrule
\end{tabular}
\end{sc}
\end{small}
\end{center}
\end{table*}

\begin{table*}[t]
\caption{Accuracy on ImageNet with various domain shifts.}
\label{table:ImageNetX}
\vspace{-10pt}
\begin{center}
\begin{small}
\begin{sc}
\begin{tabular}{lcccccc}
\toprule
 \multirow{2}*{Method} & {In-Distribution} & \multicolumn{5}{c}{Out-of-Distributions}\\
 \cmidrule(lr){2-2} \cmidrule(lr){3-7}
 & ImageNet & ImageNet-V2 & ImageNet-S & ImageNet-A & ImageNet-R & Avg. \\
\midrule
Zero-shot    &66.7 & 60.8 & 46.1 & 47.8 & 74.0 & 57.2 \\
Fine-tune    &68.2$_{\pm 0.1}$ & 61.9$_{\pm 0.1}$ & 46.8$_{\pm 0.1}$ & 46.4$_{\pm 0.1}$ & 75.1$_{\pm 0.1}$  & 57.6$_{\pm 0.1}$ \\
CoOp~(\citeyear{cite:IJCV2022CoOp})    & {71.5} & 64.2 & 48.0 & 49.7 & 75.2 & 59.3 \\
CoCoOp~(\citeyear{cite:CVPR2022CoCoOp}) & 71.0 & 64.2 & 48.8 & \textbf{50.6} & {76.2} & 59.9  \\
CLIPood & \textbf{71.6}$_{\pm 0.1}$ & \textbf{64.9}$_{\pm 0.1}$ & \textbf{49.3}$_{\pm 0.1}$ & 50.4$_{\pm 0.1}$ & \textbf{77.2}$_{\pm 0.1}$ & \textbf{60.4}$_{\pm 0.1}$ \\

\bottomrule
\end{tabular}
\end{sc}
\end{small}
\end{center}
\vspace{-5pt}
\end{table*}

\vspace{-15pt}
\section{Experiments}

\label{sec:exp}

We explore two types of out-of-distributions in this paper. One is \textit{domain shift}, where the test data \cmr{follow} different domain distributions from the training data. The other is \textit{open class}, where the test data \cmr{contain} different classes unseen in the training data. We conduct experiments on three OOD scenarios. In the first two scenarios, we explore the two OOD types separately. In the third scenario, we newly propose to solve a more general and challenging OOD situation where both domain shift and open class appear in test data. \cmr{Code is available at \href{https://github.com/thuml/CLIPood}{https://github.com/thuml/CLIPood}.}

\textbf{Implementation Details.} We use the CLIP pre-trained model with the ViT-B/16~\cite{cite:ICLR2021ViT} image encoder \cmr{and run experiments with half-precision (FP16) during training and inference}. We keep the temperature of the softmax function the same as the pre-trained model as $\tau=0.01$, and use the same hyper-parameter $\lambda=0.3$ for all datasets to avoid over-tuning on specific tasks. We adopt a batch size of $36$. We use the AdamW~\cite{cite:ICLR2019AdamW} optimizer with the cosine learning rate strategy for all datasets. By default, we set $\beta=0.5$, use a learning rate of $5\times10^{-6}$, and train for $5000$ iterations. For each result of CLIPood, we report the average result and the standard deviation of three runs with random seeds. More details can be found in supplementary materials.

\begin{table*}[ht]
\caption{Generalization performance on $11$ downstream datasets with open classes.}
\vspace{-10pt}
\label{table:openclass}
\begin{center}
\begin{sc}
\begin{small}

    \begin{subtable}[ht]{0.45\textwidth}
        \centering
        \caption{Average over 11 datasets}
        
\begin{tabular}{lccc}
\toprule
        & Base  & New   & H     \\
\midrule
CLIP            & 69.3 & 74.2 & 71.7 \\
CoOp~(\citeyear{cite:IJCV2022CoOp})           & 82.7 & 63.2 & 71.7 \\
CoCoOp~(\citeyear{cite:CVPR2022CoCoOp})          & 80.5 & 71.7 & 75.8 \\
CLIPood         & $\textbf{83.9}_{\pm 0.1}$ & $\textbf{74.5}_{\pm 0.1}$ & $\textbf{78.9}_{\pm 0.1}$ \\

\bottomrule                          
\end{tabular}
    \end{subtable}
    \hfil
    \begin{subtable}[ht]{0.45\textwidth}
        \centering
        \caption{ImageNet}
        
\begin{tabular}{lccc}
\toprule
        & Base  & New   & H     \\
\midrule
CLIP        & 72.4 & 68.1 & 70.2 \\
CoOp~(\citeyear{cite:IJCV2022CoOp})        & 76.5 & 67.9 & 71.9 \\
CoCoOp~(\citeyear{cite:CVPR2022CoCoOp})      & 76.0 & \textbf{70.4} & 73.1 \\
CLIPood         & $\textbf{77.5}_{\pm 0.1}$ & $70.3_{\pm 0.1}$ & $\textbf{73.7}_{\pm 0.1}$ \\
\bottomrule                          
\end{tabular}
    \end{subtable}
    
\end{small}
\end{sc}
\end{center}
\vspace{-5pt}
\end{table*}

\begin{table*}[t]
\setlength{\tabcolsep}{3.6pt}
\caption{Accuracy on OfficeHome and DomainNet with both domain shift and open classes.}
\label{table:ODG}
\vspace{-10pt}
\begin{center}
\begin{small}
\begin{sc}
\begin{tabular}{llcccccccccc}
\toprule
 \multirow{2}*{Split} & \multirow{2}*{Method} & \multicolumn{4}{c}{OfficeHome} & \multicolumn{6}{c}{DomainNet}\\
 \cmidrule(lr){3-6} \cmidrule(lr){7-12} &  & A & C & P & R & C & I & P & Q & R & S \\

\midrule

\multirow{3}*{\cmr{Base}} & CLIP    &  86.8 & 75.5 & 89.5 & 92.6 & 72.8 & 51.7 & 66.0 & 13.5 & 83.4 & 66.9\\
& CoOp  & 87.0$_{\pm 0.4}$ & 78.3$_{\pm 1.2}$ & 92.4$_{\pm 0.2}$ & 91.4$_{\pm 0.6}$ & 75.7$_{\pm 0.2}$ & 58.8$_{\pm 0.5}$ & 68.5$_{\pm 1.3}$ & 13.1$_{\pm 1.0}$ & 84.0$_{\pm 0.5}$ & 70.0$_{\pm 0.1}$ \\
& CLIPood & \textbf{90.1}$_{\pm 0.2}$ & \textbf{79.7}$_{\pm 0.2}$ & \textbf{93.1}$_{\pm 0.1}$ & \textbf{94.8}$_{\pm 0.1}$ & \textbf{79.0}$_{\pm 0.2}$ & \textbf{62.2}$_{\pm 0.1}$ & \textbf{73.0}$_{\pm 0.2}$ & \textbf{20.2}$_{\pm 0.2}$ & \textbf{86.2}$_{\pm 0.1}$ & \textbf{73.8}$_{\pm 0.1}$ \\

\midrule

\multirow{3}*{\cmr{New}} & CLIP    &  76.6 & 59.4 & 88.1 & 86.2 & 70.2 & 44.1 & 66.4 & 14.1 & 83.5 & 61.0\\
& CoOp  & 76.5$_{\pm 1.1}$ & 56.6$_{\pm 2.4}$ & 88.0$_{\pm 1.9}$ & \textbf{86.8}$_{\pm 0.7}$ & \textbf{71.5}$_{\pm 0.2}$ & 47.2$_{\pm 0.3}$ & 67.3$_{\pm 0.7}$ & 14.8$_{\pm 0.7}$ & \textbf{83.7}$_{\pm 0.7}$ & \textbf{63.1}$_{\pm 0.3}$ \\
& CLIPood & \textbf{77.8}$_{\pm 0.2}$ & \textbf{60.0}$_{\pm 0.2}$ & \textbf{88.3}$_{\pm 0.1}$ & 86.7$_{\pm 0.1}$ & 71.2$_{\pm 0.1}$ & \textbf{48.1}$_{\pm 0.1}$ & \textbf{68.2}$_{\pm 0.2}$ & \textbf{18.0}$_{\pm 0.4}$ & 83.4$_{\pm 0.1}$ & 62.9$_{\pm 0.1}$ \\

\midrule
\multirow{3}*{Total} & CLIP    &  82.6 & 67.3 & 88.8 & 89.5 & 71.4 & 47.1 & 66.2 & 13.8 & 83.4 & 63.4\\
& CoOp  & 82.7$_{\pm 0.5}$ & 67.2$_{\pm 0.7}$ & 90.2$_{\pm 1.0}$ & 89.2$_{\pm 0.6}$ & 73.4$_{\pm 0.3}$ & 51.8$_{\pm 0.3}$ & 67.9$_{\pm 1.0}$ & 13.7$_{\pm 0.8}$ & 83.9$_{\pm 0.5}$ & 66.0$_{\pm 0.2}$ \\
& CLIPood & \textbf{85.1}$_{\pm 0.1}$ & \textbf{69.6}$_{\pm 0.2}$ & \textbf{90.8}$_{\pm 0.1}$ & \textbf{91.0}$_{\pm 0.1}$ & \textbf{74.8}$_{\pm 0.1}$ & \textbf{53.6}$_{\pm 0.1}$ & \textbf{70.6}$_{\pm 0.1}$ & \textbf{19.1}$_{\pm 0.3}$ & \textbf{84.8}$_{\pm 0.1}$ & \textbf{67.4}$_{\pm 0.1}$ \\

\bottomrule
\end{tabular}
\end{sc}
\end{small}
\end{center}
\vspace{-5pt}
\end{table*}

\subsection{Generalize CLIP to Domain Shift}\label{sec:DG}

\textbf{Benchmarks.} We evaluate generalization to domain shift with two benchmarks. On the first benchmark, we use five multi-domain datasets in DomainBed~\cite{cite:ICLR2021DomainBed}: PACS~\cite{cite:ICCV2017PACS}, VLCS~\cite{cite:CVPR2021VLCS}, OfficeHome~\cite{cite:CVPR2017OfficeHome}, TerraIncognita~\cite{cite:ECCV2018Terra} and DomainNet~\cite{cite:ICCV2019DomainNet}. We follow the train-validate-test split of each dataset as the DomainBed benchmark and the leave-one-out evaluation protocol, where at each time, one domain is chosen as the test domain for evaluating OOD generalization, and other domains are chosen as the training domains.

On the second benchmark, we use ImageNet~\cite{cite:CVPR2009ImageNet} as the training dataset and evaluate the performance on four variants of ImageNet with distribution shifts: ImageNet-V2~\cite{cite:ICML2019ImageNetV2}, ImageNet-Sketch~\cite{cite:NIPS2019ImageNetSketch}, ImageNet-A~\cite{cite:CVPR2021ImageNetA} and ImageNet-R~\cite{cite:ICCV2021ImageNetR}. We follow the protocol in~\cite{cite:CVPR2022CoCoOp} and randomly sample a $16$-shot training set while using the original test set for evaluation.

\textbf{Results.} For each dataset in the DomainBed benchmark, we report the average accuracy on all test domains in Table~\ref{table:domainbed}. We consider the methods with the CLIP pre-trained model and the ResNet-50~\cite{cite:CVPR2016ResNet} model pre-trained on ImageNet. We compare with the zero-shot performance and standard fine-tuning of the model using all training domains (ERM). \cmr{For ERM with CLIP, we follow \citet{cite:CVPR2022Wise} and initialize the classifier head with text embeddings to achieve competitive performance. Its results are reported from our own implementation, following the details mentioned in our paper.} We also compare with the state-of-the-arts using the CLIP pre-trained model for domain generalization: MIRO~\cite{cite:Arxiv2022MIRO}, DPL~\cite{cite:Arxiv2022DomainPrompt}, and the best-performing method using ResNet-50 reported in DomainBed~\cite{cite:ICLR2021DomainBed}, CORAL~\cite{cite:ECCV2016DeepCoral}. \cmr{We present the results reported in original papers of these methods for comparison in Table~\ref{table:domainbed}.} \cmr{For some methods such as MIRO and other methods where the original papers do not report results under our setting, we also re-implement them with our desings for a unified comparison, and these results are shown in Appendix \ref{apdx:DGbaselines}.} CLIPood outperforms methods with ResNet pre-trained model by a large margin, which indicates that utilizing knowledge in vision-language models provides a promising way for improving OOD generalization. It also outperforms state-of-the-arts using CLIP models: MIRO and DPL, indicating CLIPood is a simple and effective method to generalize CLIP to out-of-distributions. We further report the OOD generalization results on different variants of ImageNet in Table~\ref{table:ImageNetX}. CLIPood achieves comparable performance on the in-distribution test data and outperforms the state-of-the-art methods CoOp~\cite{cite:IJCV2022CoOp} and CoCoOp~\cite{cite:CVPR2022CoCoOp} on the OOD datasets, which shows its generalizability on various domain shifts.

\subsection{Generalize CLIP to Open Class}\label{sec:OOC}

\textbf{Benchmarks.} We evaluate generalization to open classes on the benchmark covering a diverse set of recognition tasks, including general object classification: ImageNet~\cite{cite:CVPR2009ImageNet} and Caltech101~\cite{cite:CVPRW2004Caltech}; fine-grained classification: OxfordPets~\cite{cite:CVPR2012OxfordPets}, StanfordCars~\cite{cite:ICCVW2013StanfordCars}, Flowers102~\cite{cite:ICVGIP2008Flowers}, Food101~\cite{cite:ECCV2014Food} and FGVCAircraft~\cite{cite:Arxiv2013Aircraft}; specific classification tasks: SUN397~\cite{cite:CVPR2010SUN} for scene recognition, UCF101~\cite{cite:Arxiv2012UCF101} for action recognition, DTD~\cite{cite:CVPR2014DTD} for texture classification and EuroSAT~\cite{cite:EuroSAT} for satellite image recognition. We follow the protocol in ~\cite{cite:CVPR2022CoCoOp} and split the classes in each dataset equally into two parts, one as base classes and the other as new classes. We train the model on base-class data and test on base classes and new classes separately to evaluate the generalization ability.

\textbf{Results.} The results of generalizing CLIP to open classes are shown in Table~\ref{table:openclass}. We report the accuracies on base classes and new classes as well as their harmonic mean (H) to highlight the trade-off between downstream adaptation and open-class generalization. We compare CLIPood with the zero-shot prediction of CLIP and the state-of-the-art methods on this benchmark: CoOp~\cite{cite:IJCV2022CoOp} and CoCoOp~\cite{cite:CVPR2022CoCoOp}. The detailed results on each dataset are presented in the appendix. As shown in the table, CoOp suffers from a large decrease in the accuracy of new classes after its adaptation to base classes. 
CoCoOp mitigates the decrease in the accuracy of new classes but sacrifices the adaptation performance on the base classes, and the gap with zero-shot performance on new classes is still large. On some datasets such as ImageNet, CoCoOp and CLIPood successfully improve the performance on unseen classes over zero-shot prediction by adapting the model with related training classes. Comparing the average results, CLIPood outperforms zero-shot prediction and existing methods by a large margin, showing that it simultaneously adapts the model to improve the performance on the downstream tasks and keeps the OOD generalization ability on open classes.

\begin{figure*}[htbp]
    \centering
    \subfloat[Ablation Study]{\label{fig:analysis_ablation}\includegraphics[width=.24\textwidth]{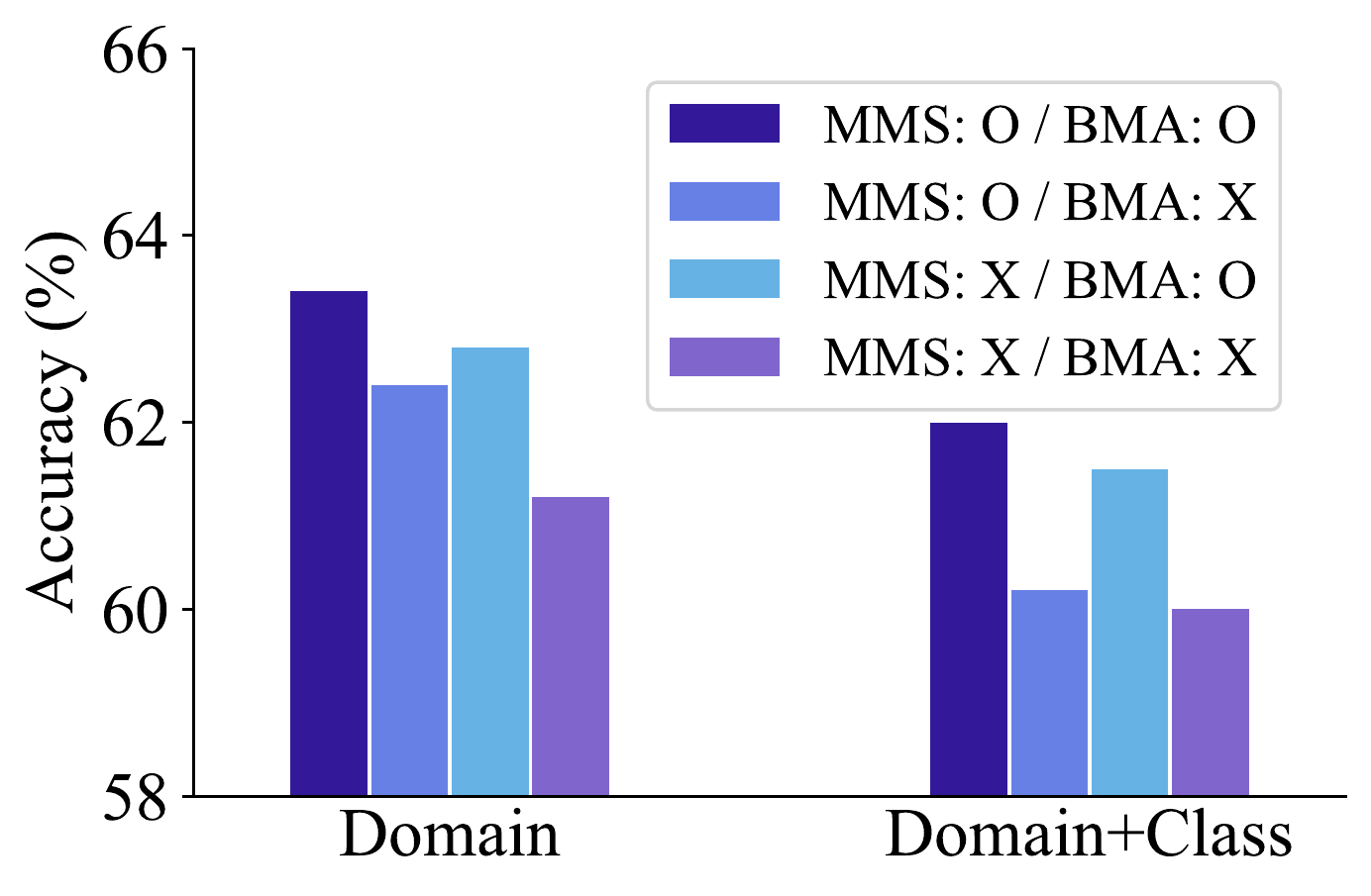}}
    \subfloat[Metric Softmax]{\label{fig:analysis_metric}\includegraphics[width=.24\textwidth]{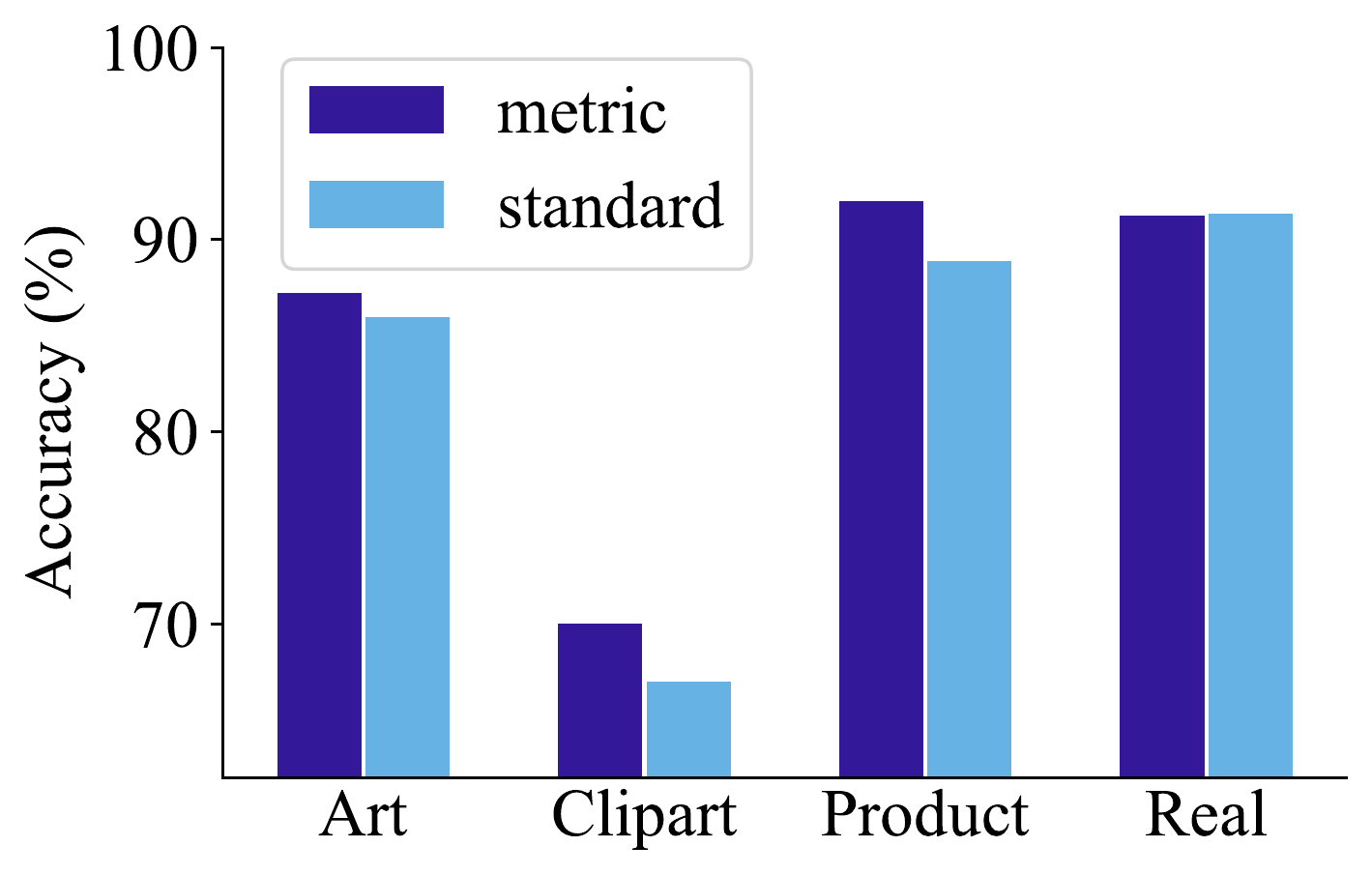}}
    \subfloat[Adaptive Margin]{\label{fig:analysis_margin}\includegraphics[width=.24\textwidth]{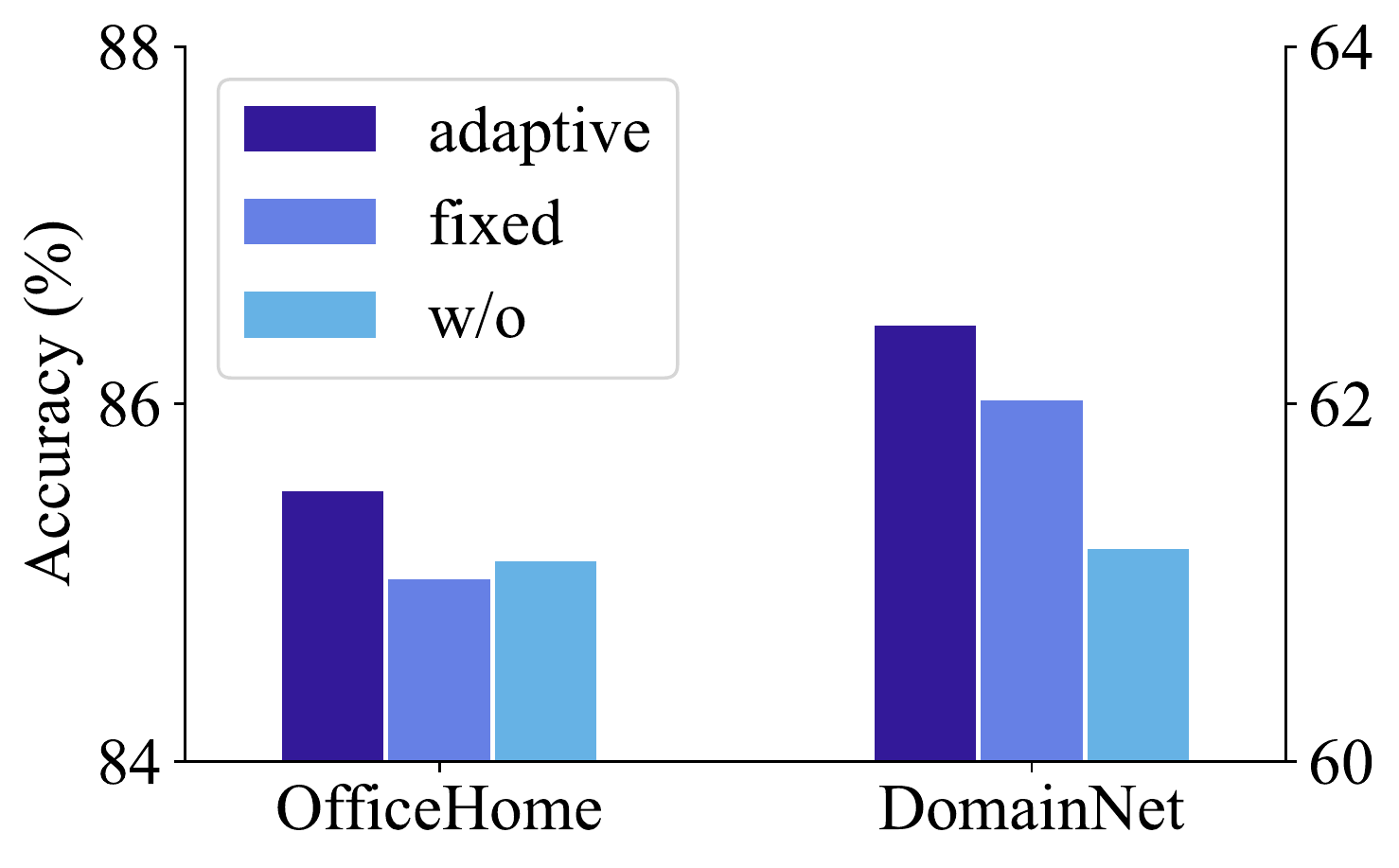}}
    \subfloat[Beta Moving Average]{\label{fig:analysis_bma}\includegraphics[width=.24\textwidth]{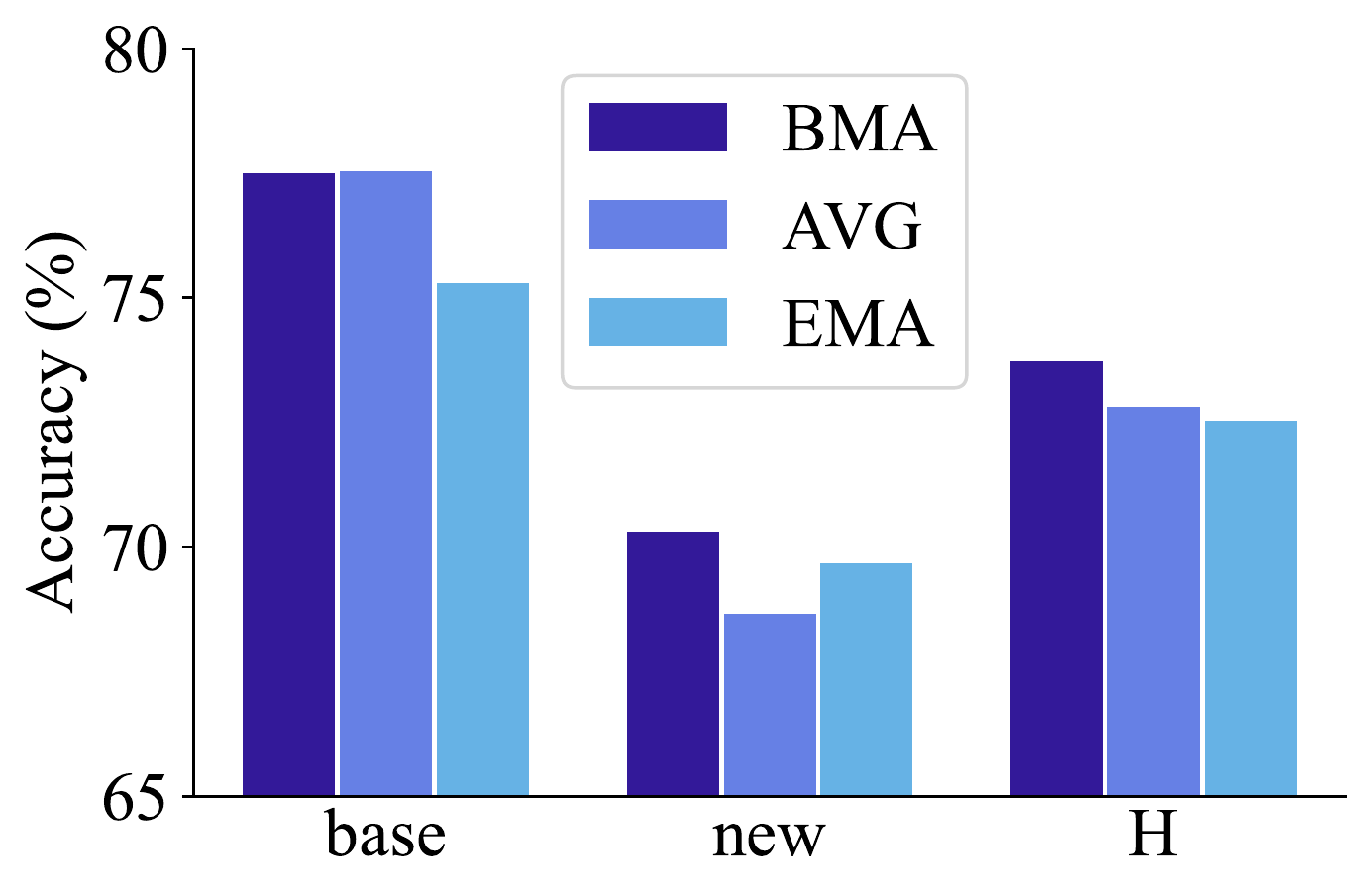}}
    
    \caption{Analysis experiments for CLIPood.}
    \label{fig:analysis}
\end{figure*}

\begin{figure*}[htbp]
\begin{center}
\centerline{\includegraphics[width=0.95\textwidth]{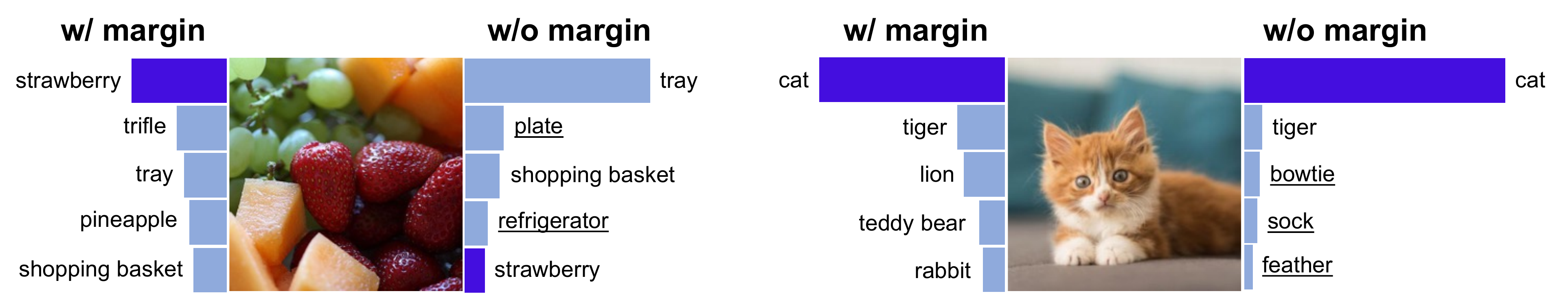}}
\vspace{-5pt}
\caption{Predictions from models trained with and without adaptive margin.}
\label{fig:mms_visualize}
\vspace{-25pt}
\end{center}
\end{figure*}

\subsection{Generalize CLIP to Domain Shift and Open Class}\label{sec:ODG}

\textbf{Benchmarks.} We further propose a more realistic in-the-wild setting where both domain shift and open class may appear in the test data. We choose OfficeHome and DomainNet from DomainBed because they have sufficient numbers of classes for evaluating open class situations. We split the classes in each dataset into two parts, one as base classes and the other as new classes. We adopt the leave-one-domain-out protocol and 
train the model on base-class data in training domains and test on all test data with both base and new classes to evaluate the OOD generalization ability. We put the text embeddings of base and new classes together for the classification of all test data, which is a more realistic OOD testing protocol, since we cannot know whether the test data is from base or new classes in advance. 

\textbf{Results.} The experimental results are shown in Table~\ref{table:ODG}. We report the accuracy of each test domain. We compare with zero-shot performance (CLIP) and CoOp and do not reimplement CoCoOp because of its great memory and time cost on these datasets. CoOp improves upon zero-shot on some domains but also suffers from degradation in other domains. CLIPood consistently outperforms zero-shot and CoOp on all test domains, showing its effectiveness in more general and realistic OOD situations.

\subsection{Analysis of CLIPood}

\textbf{Ablation Study.} We explore the efficacy of each module in CLIPood, including Margin Metric Softmax (MMS) and Beta Moving Average (BMA). We compare CLIPood with its variants with or without MMS and BMA on the DomainNet dataset. We report the average results over all test domains with domain shift (Domain) and under the OOD setting with both domain shift and open classes (Domain+Class). From Figure~\ref{fig:analysis_ablation}, we can observe that adding MMS and BMA improves the generalization performance on the domain shift and open class situations, which demonstrates the effectiveness of these two modules. CLIPood incorporates both these two modules and achieves the best performance, which demonstrates that the designs from the aspects of training objective and model optimization work together towards better OOD generalization of CLIP. Note that the variant without MMS and BMA still outperforms standard fine-tuning because it utilizes metric softmax as the training objective, which we will also discuss below.

\textbf{Analysis on CLIP Fine-tuning.} In CLIPood, we choose to fine-tune the model by metric softmax, which compares the image and class text embeddings. Here we compare it with standard fine-tuning, which adds and trains a parametric linear classifier together with the pre-trained backbone. We show the performance on OfficeHome with domain shifts as in Section~\ref{sec:DG}. As shown in Figure~\ref{fig:analysis_metric}, metric softmax fine-tuning outperforms standard fine-tuning on most of the test domains, which indicates that metric softmax is a better choice to improve OOD generalization for CLIP.

\begin{table*}[ht]
\centering
\caption{Performance with different weight ensemble methods.}
\label{table:ensemble}
\begin{small}
\begin{sc}
\begin{tabular}{lccccccc}
\toprule
\multirow{2}*{Method}                 & \multirow{2}*{DomainNet}     & \multicolumn{3}{c}{ImageNet} & \multicolumn{3}{c}{SUN397} \\ \cmidrule(lr){3-5} \cmidrule(lr){6-8}
& & Base & New & H & Base & New & H \\
\midrule
CLIPood w/ SWAD        & 62.3          & 77.2          & 68.7         & 72.7 & 80.7	& 78.2	& 79.4       \\
CLIPood w/ WiSE-FT     & 62.7          & 77.3          & 69.9         & 73.4  & 79.2	& 78.8	& 79.0     \\
CLIPood w/ BMA (Ours) & \textbf{63.5}          & \textbf{77.5}          & \textbf{70.3}         & \textbf{73.7}  & \textbf{81.0}	& \textbf{79.3}	& \textbf{80.2}     \\
\bottomrule
\end{tabular}
\end{sc}
\end{small}
\end{table*}

\textbf{Analysis on Adaptive Margin.} Margin Metric Softmax (MMS) adds an adaptive margin to preserve inherent unequal relations of classes in the language space during fine-tuning. We compare MMS to the variants with a fixed margin for all data or without margin on OfficeHome and DomainNet with both domain shift and open classes as in Section~\ref{sec:ODG}. \cmr{As shown in Figure~\ref{fig:analysis_margin}, adding a margin in metric softmax may improve the performance. Still, a fixed margin does not give a stable improvement on the variant without margins (the vanilla metric softmax), and an adaptive margin consistently outperforms the variants with a fixed margin or no margins, especially achieving remarkable (1.2\%) improvement on DomainNet. This demonstrates the efficacy of the proposed Margin Metric Softmax for OOD generalization of the CLIP model.}

In Figure~\ref{fig:mms_visualize}, we show the top-$5$ predictions on test images from the models trained with and without adaptive margins. The \texttt{cat} image comes from a domain with the distribution shift. The model without margins outputs unrelated classes (underlined) such as \texttt{bowtie} (probably because the ears of the \texttt{cat} may look like a \texttt{bowtie}), indicating that it may overfit the image modality and forget semantic relations. The model trained with margins outputs related classes of animals. The \texttt{strawberry} image comes from a new class unseen during training. The model without margins outputs unrelated classes and makes wrong predictions, while the model with margins makes related and right predictions. These results show that the adaptive margin keeps the semantic relationship of classes better and improves generalization under domain shift and open classes.

\textbf{Analysis on BMA.} To evaluate the efficacy of the proposed Beta Moving Average (BMA), we compare it with the commonly used Exponential Moving Average (EMA) and the uniform average (AVG) of the model checkpoints. We report results on ImageNet with open classes as in Section~\ref{sec:OOC}. As shown in Figure~\ref{fig:analysis_bma}, EMA focuses on only one side of the training trajectory, which may perform worse than average weighting. BMA performs the best balance on base and new classes, showing the importance of the Beta-distribution weighting to focus on both zero-shot and fine-tuned models. 

\cmr{We also compare BMA with other weight ensemble methods. We re-implement and compare with SWAD~\cite{cite:NIPS2021SWAD} and WiSE-FT~\cite{cite:CVPR2022Wise}. We investigate how CLIPood performs when BMA in our method is replaced with SWAD or WiSE-FT, which are denoted as {CLIPood w/ SWAD} or CLIPood {w/ WiSE-FT}. We conduct experiments on DomainNet for domain shift settings and on ImageNet and SUN397 for open class settings. As shown in Table \ref{table:ensemble}, for both domain shift and open class, CLIPood with BMA outperforms CLIPood with SWAD and WiSE-FT in most cases, which also shows that BMA is a better model averaging choice for CLIP models.}

\section{Discussion}
\label{apdx:limitation}
\cmr{
CLIPood consistently outperforms zero-shot pre-trained models and existing generalization techniques of pre-trained models, showing its efficacy for different OOD generalization situations. Here we also discuss some limitations of this method and some possible future directions regarding these limitations. In CLIPood, we propose Metric Margin Softmax and Beta Moving Average, which introduce negligible additional costs in storage or computation compared with standard fine-tuning. Computing BMA requires the storage of one more model during training. It would be imperceptible in common situations but may be constrained when the memory resource is extremely limited. One of the limitations is that we mainly consider better fine-tuning the image encoder for OOD generalization. A possible future work regarding this would be exploring adaptation on both text and image modalities for better OOD generalization. Another limitation is that the performance of our method may still be influenced by the zero-shot performance of the pre-trained model. For some domains or classes with extremely poor zero-shot performance, indicating that the pre-trained knowledge may not help certain cases, the improvements from exploiting knowledge in pre-trained models may be minor. A possible future work regarding this may need to consider the whole pre-training-fine-tuning pipeline, exploring pre-trained models with better generalization and designing corresponding adaptation methods for them.
}

\section{Conclusion}

In this paper, we propose to solve the problem of generalizing CLIP to out-of-distributions with both domain shift and open classes in downstream tasks. We propose CLIPood to fine-tune CLIP with a simple and effective design to improve its OOD generalization. CLIPood introduces Margin Metric Softmax, which adds class adaptive margins in metric softmax training to exploit semantic relations between classes from the text modality. It also introduces Beta Moving Average to maintain a temporal ensemble according to a Beta distribution, which incorporates both the zero-shot model and the adapted model. Experiments on various datasets with different OOD scenarios show that CLIPood consistently outperforms existing generalization techniques.

\section*{Acknowledgements}
This work was supported by the National Key Research and Development Plan (2020AAA0109201), National Natural Science Foundation of China (62022050 and 62021002), and Beijing Nova Program (Z201100006820041).

\bibliography{example_paper}
\bibliographystyle{icml2023}

\newpage
\appendix
\onecolumn
\section{Experimental Details}
\label{apdx:implementation}

\subsection{Implementation Details}
We use the CLIP pre-trained model with ViT-B/16~\cite{cite:ICLR2021ViT} as the image encoder. We keep the temperature of the softmax function the same as the pre-trained model as $\tau=0.01$. All the default configs are shown in Table \ref{apdx:default_cfg}. We use the same hyper-parameter $\lambda=0.3$ for all datasets to avoid over-tuning on specific tasks. We adopt a batch size of $36$ except for DomainNet with $60$, and all images are randomly resized and cropped to $224\times224$. We use the AdamW~\cite{cite:ICLR2019AdamW} optimizer with the cosine learning rate strategy for all datasets. By default, we set $\beta=0.5$, and use a learning rate of $5\times10^{-6}$ and train for $5000$ iterations. Considering the small amount of data, we set $\beta$ to $0.1$ for all 16-shot benchmarks. For specific datasets, we adjust different numbers of iterations and learning rates. We train $10000$ iterations for DomainNet, $2500$ iterations for StanfordCars and SUN397, $1000$ iterations for UCF101, and $500$ iterations for other 16-shot datasets except for ImageNet and FGVCAirCraft. We adopt the learning rate of $1\times10^{-5}$ for DomainNet and OfficeHome in the case of domain shifts. For each result of CLIPood, we report the average result and the standard deviation of three runs with random seeds.

\begin{table*}[ht]
\centering
\caption{Default configs for the experiments.}
\label{apdx:default_cfg}
\begin{tabular}{l|c}
\toprule
Default Config    & Value       \\
\midrule
optimizer         & AdamW   \\
base lr           & $5\times10^{-6}$    \\
weight decay      & 0.1 \\
lr scheduler      & cosine decay  \\
augmentation      & \texttt{RandomResizedCrop}   \\
batch size        & 36 \\
\# iterations     & 5000 \\
temperature       & 0.01 \\
$\lambda$ for MMS & 0.3 \\
$\beta$ for BMA   & 0.5\\
\bottomrule
\end{tabular}
\vskip 0.1in
\end{table*}

\subsection{Prompt Templates for Each Dataset}
By default, we use ``a photo of \texttt{[CLASS]}.'' as the prompt template for class labels, where \texttt{[CLASS]} refers to the name of a class with the hyphens replaced by spaces. Following \citet{cite:IJCV2022CoOp}, for fine-grained classification datasets such as FGVCAircraft, we add the name of the superclass (aircraft) or the description to the template. The full templates for all datasets are shown as follows.
\begin{itemize}
    \item OxfordPets: ``a photo of a \texttt{[CLASS]}, a type of pet.''
    \item FGVCAircraft: ``a photo of a \texttt{[CLASS]}, a type of aircraft.''
    \item DTD: ``\texttt{[CLASS]} texture.''
    \item EuroSAT: ``a centered satellite photo of \texttt{[CLASS]}.''
    \item Food101: ``a photo of a \texttt{[CLASS]}, a type of food.''
    \item UCF101: ``a photo of a person doing \texttt{[CLASS]}.''
    \item Other datasets: ``a photo of a \texttt{[CLASS]}.''
\end{itemize}

\subsection{Computing Infrastructure}

For the experiments, we use PyTorch 1.13.1, torchvision 0.14.1, and CUDA 11.6 libraries. We use a machine with 32 CPUs, 256 GB memory, and the NVIDIA TITAN X GPU.

\subsection{Licenses of Datasets}
\label{apdx:license}
OxfordPets is under CC BY-SA 4.0 license. Other datasets are publicly available online and under custom licenses for non-commercial usage.

\section{More Experimental Results}

\subsection{\cmr{Unified Comparison with More Methods on Domain Shifts}}
\label{apdx:DGbaselines}

We compare our method with other state-of-the-art baselines, including WiSE-FT~\cite{cite:CVPR2022Wise}, FLYP~\cite{cite:Arxiv2023FLYP}, LP-FT~\cite{cite:Arxiv2022LPFT}, and MIRO~\cite{cite:Arxiv2022MIRO} with SWAD~\cite{cite:NIPS2021SWAD}. Since there are no available results in our setting, we re-implement them with our codebase for a unified comparison. We report the accuracy on each domain of DomainNet. Results are shown in Table \ref{table:DGbaselines}.
\cmr{Under our setting, all the state-of-the-art methods give better performance than ERM. Among them, MIRO benefits from SWAD, and MIRO+SWAD achieves previous state-of-the-art results. Compared with these baselines, CLIPood can outperform them generally with a significant accuracy gain.}

\begin{table}[ht]
\centering
\caption{Performance of more baselines on DomainNet.}
\label{table:DGbaselines}
\begin{sc}
\begin{small}
\begin{tabular}{lccccccc}
\toprule
Method         & Clipart       & Infograph     & Painting      & Quickdraw     & Real          & Sketch        & Average       \\
\midrule
ERM            & 76.3          & 47.8          & 68.1          & 19.7          & 80.9          & 66.5          & 59.9          \\
WiSE-FT        & 76.8          & 49.5          & 69.4          & 20.1          & 81.7          & 67.2          & 60.8          \\
FLYP+WiSE-FT   & \textbf{78.1} & 51.9          & 69.8          & 20.0          & 84.3          & 67.6          & 61.9          \\
LP-FT          & 75.1          & 51.7          & 70.7          & 16.2          & 85.0          & 67.1          & 61.0          \\
MIRO           & 76.6          & 51.0          & 70.9          & 18.8          & 82.3          & 68.5          & 61.3          \\
MIRO+SWAD      & 75.9          & 51.6          & 71.3          & 20.0          & 82.5          & 68.4          & 61.6          \\
CLIPood (Ours) & 77.6          & \textbf{54.7} & \textbf{72.5} & \textbf{20.7} & \textbf{85.7} & \textbf{69.9} & \textbf{63.5} \\
\bottomrule
\end{tabular}
\end{small}
\end{sc}
\end{table}

\subsection{Detailed Results on Open Classes}

We report the full results of generalizing to open classes as in Section~\ref{sec:OOC}. We report the results in each dataset and the average results over all $11$ datasets in Table \ref{table:sup_openclass}. On most of the datasets, CLIPood achieves better adaptation performance on base classes and still narrows the gap between zero-shot prediction on new classes or even performs better. Comparing the average results, CLIPood outperforms zero-shot prediction and existing methods by a large margin, showing that it simultaneously adapts the model to improve the performance on the downstream tasks and keeps the OOD generalization ability on open classes.

\begin{figure}[ht]
    \centering
    \includegraphics[width=0.4\textwidth]{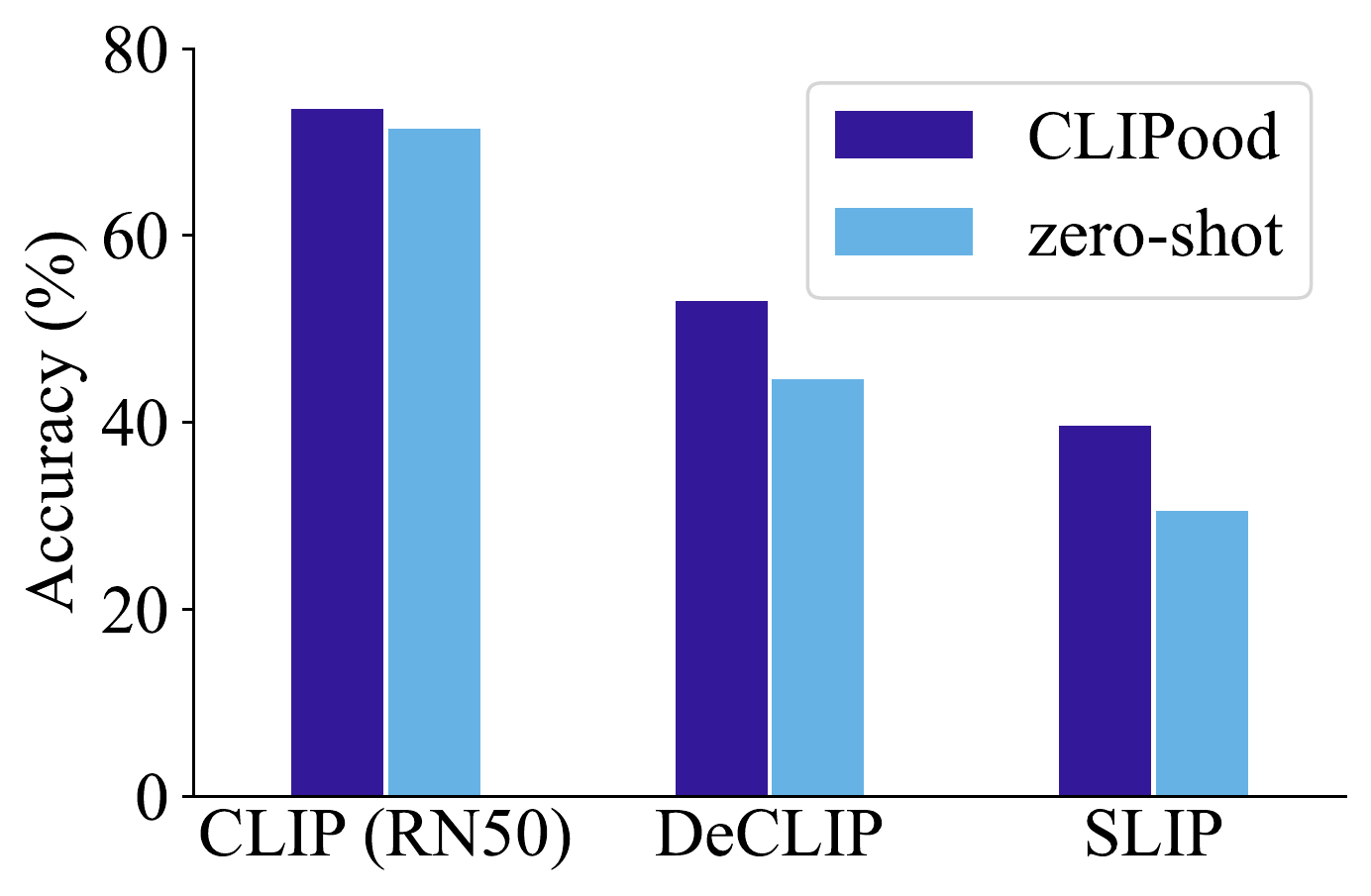}
    \caption{Results on other vision-language models.}
    \label{fig:backbone}
\end{figure}

\begin{table*}[tbp]
\caption{Generalization performance on $11$ downstream datasets with open classes.}
\label{table:sup_openclass}
\vspace{-5pt}
\begin{center}
\begin{sc}
\begin{small}

    \begin{subtable}[ht]{0.48\textwidth}
        \centering
        \caption{Average over 11 datasets}
        \vspace{-5pt}
        
\begin{tabular}{lccc}
\toprule
        & Base  & New   & H     \\
\midrule
CLIP            & 69.3 & 74.2 & 71.7 \\
CoOp           & 82.7 & 63.2 & 71.7 \\
CoCoOp          & 80.5 & 71.7 & 75.8 \\
CLIPood         & $\textbf{83.9}_{\pm 0.1}$ & $\textbf{74.5}_{\pm 0.1}$ & $\textbf{78.9}_{\pm 0.1}$ \\

\bottomrule                          
\end{tabular}
    \end{subtable}
    \hfill
    \begin{subtable}[ht]{0.48\textwidth}
        \centering
        \caption{ImageNet}
        \vspace{-5pt}
        
\begin{tabular}{lccc}
\toprule
        & Base  & New   & H     \\
\midrule
CLIP        & 72.4 & 68.1 & 70.2 \\
CoOp        & 76.5 & 67.9 & 71.9 \\
CoCoOp      & 76.0 & \textbf{70.4} & 73.1 \\
CLIPood         & $\textbf{77.5}_{\pm 0.1}$ & $70.3_{\pm 0.1}$ & $\textbf{73.7}_{\pm 0.1}$ \\
\bottomrule                          
\end{tabular}
    \end{subtable}
    
    \vskip 0.1in
    \begin{subtable}[ht]{0.48\textwidth}
        \centering
        \caption{Caltech101}
        \vspace{-5pt}
        
\begin{tabular}{lccc}
\toprule
        & Base  & New   & H     \\
\midrule
CLIP        & 96.8 & 94.0 & 95.4 \\
CoOp        & 98.0 & 89.8 & 93.7 \\
CoCoOp      & 98.0 & 93.8 & 95.8 \\
CLIPood     & $\textbf{98.7}_{\pm 0.1}$	& $\textbf{94.6}_{\pm 0.1}$	& $\textbf{96.6}_{\pm 0.1}$ \\
\bottomrule                          
\end{tabular}
    \end{subtable}
    \hfill
    \begin{subtable}[ht]{0.48\textwidth}
        \centering
        \caption{OxfordPets}
        \vspace{-5pt}
        
\begin{tabular}{lccc}
\toprule
        & Base  & New   & H     \\
\midrule
CLIP        & 91.2 & 97.3 & 94.1 \\
CoOp        & 93.7 & 95.3 & 94.5 \\
CoCoOp      & 95.2 & \textbf{97.7} & \textbf{96.4} \\
CLIPood         & $\textbf{95.7}_{\pm 0.1}$ & $96.4_{\pm 0.2}$ & $96.0_{\pm 0.1}$ \\
\bottomrule                          
\end{tabular}
    \end{subtable}
    
    \vskip 0.1in
    \begin{subtable}[ht]{0.48\textwidth}
        \centering
        \caption{StanfordCars}
        \vspace{-5pt}
        \begin{tabular}{lccc}
\toprule
        & Base  & New   & H     \\
\midrule
CLIP        & 63.4 & \textbf{74.9} & 68.7 \\
CoOp        & 78.1 & 60.4 & 68.1 \\
CoCoOp      & 70.5 & 73.6 & 72.0 \\
CLIPood         & $\textbf{78.6}_{\pm 0.1}$ & $73.5_{\pm 0.3}$ & $\textbf{75.9}_{\pm 0.2}$ \\
\bottomrule                          
\end{tabular}
    \end{subtable}
    \hfill
    \begin{subtable}[ht]{0.48\textwidth}
        \centering
        \caption{Flowers102}
        \vspace{-5pt}
        \begin{tabular}{lccc}
\toprule
        & Base  & New   & H     \\
\midrule
CLIP        & 72.1 & \textbf{77.8} & 74.8 \\
CoOp        & \textbf{97.6} & 59.7 & 74.1 \\
CoCoOp      & 94.9 & 71.8 & 81.7 \\
CLIPood         & $93.5_{\pm 0.2}$ & $74.5_{\pm 0.5}$ & $\textbf{82.9}_{\pm 0.3}$ \\
\bottomrule                          
\end{tabular}
    \end{subtable}

    \vskip 0.1in
    \begin{subtable}[ht]{0.48\textwidth}
        \centering
        \caption{Food101}
        \vspace{-5pt}
        
\begin{tabular}{lccc}
\toprule
        & Base  & New   & H     \\
\midrule
CLIP        & 90.1 & 91.2 & 90.7 \\
CoOp        & 88.3 & 82.3 & 85.2 \\
CoCoOp      & \textbf{90.7} & 91.3 & 91.0 \\
CLIPood         & $\textbf{90.7}_{\pm 0.1}$ & $\textbf{91.7}_{\pm 0.1}$ & $\textbf{91.2}_{\pm 0.1}$ \\
\bottomrule                          
\end{tabular}

    \end{subtable}
    \hfill
    \begin{subtable}[ht]{0.48\textwidth}
        \centering
        \caption{FGVCAircraft}
        \vspace{-5pt}
        
\begin{tabular}{lccc}
\toprule
        & Base  & New   & H     \\
\midrule
CLIP        & 27.2 & 36.3 & 31.1 \\
CoOp        & 40.4 & 22.3 & 28.8 \\
CoCoOp      & 33.4 & 23.7 & 27.7 \\
CLIPood         & $\textbf{43.3}_{\pm 0.3}$	& $\textbf{37.2}_{\pm 0.5}$ &	$\textbf{40.0}_{\pm 0.4}$ \\
\bottomrule                          
\end{tabular}
    \end{subtable}
    
    \vskip 0.1in
    \begin{subtable}[ht]{0.48\textwidth}
        \centering
        \caption{SUN397}
        \vspace{-5pt}
        \begin{tabular}{lccc}
\toprule
        & Base  & New   & H     \\
\midrule
CLIP        & 69.4 & 75.4 & 72.2 \\
CoOp        & 80.6 & 65.9 & 72.5 \\
CoCoOp      & 79.7 & 76.9 & 78.3 \\
CLIPood         & $\textbf{81.0}_{\pm 0.1}$ & $\textbf{79.3}_{\pm 0.1}$ & $\textbf{80.2}_{\pm 0.1}$ \\
\bottomrule                          
\end{tabular}
    \end{subtable}
    \hfill
    \begin{subtable}[ht]{0.48\textwidth}
        \centering
        \caption{DTD}
        \vspace{-5pt}
        
\begin{tabular}{lccc}
\toprule
        & Base  & New   & H     \\
\midrule
CLIP        & 53.2 & \textbf{59.9} & 56.4 \\
CoOp        & 79.4 & 41.2 & 54.2 \\
CoCoOp      & 77.0 & 56.0 & 64.9 \\
CLIPood         & $\textbf{80.8}_{\pm 0.6}$ & $58.6_{\pm 0.6}$ & $\textbf{67.9}_{\pm 0.3}$ \\
\bottomrule                          
\end{tabular}
    \end{subtable}
    
    \vskip 0.1in
    \begin{subtable}[ht]{0.48\textwidth}
        \centering
        \caption{EuroSAT}
        \vspace{-5pt}
        
\begin{tabular}{lccc}
\toprule
        & Base  & New   & H     \\
\midrule
CLIP        & 56.5 & \textbf{64.1} & 60.0 \\
CoOp        & 92.2 & 54.7 & 68.7 \\
CoCoOp      & 87.5 & 60.0 & 71.2 \\
CLIPood     & $\textbf{97.5}_{\pm 0.2}$ & $\textbf{64.1}_{\pm 1.1}$ & $\textbf{77.3}_{\pm 0.8}$ \\
\bottomrule                          
\end{tabular}
    \end{subtable}
    \hfill
    \begin{subtable}[ht]{0.48\textwidth}
        \centering
        \caption{UCF101}
        \vspace{-5pt}
        
\begin{tabular}{lccc}
\toprule
        & Base  & New   & H     \\
\midrule
CLIP        & 70.5 & 77.5 & 73.9 \\
CoOp        & 84.7 & 56.1 & 67.5 \\
CoCoOp      & 82.3 & 73.5 & 77.6 \\
CLIPood         & $\textbf{85.7}_{\pm 0.1}$ & $\textbf{79.3}_{\pm 0.2}$ & $\textbf{82.4}_{\pm 0.1}$ \\
\bottomrule                          
\end{tabular}
    \end{subtable}
    
\end{small}
\end{sc}
\end{center}
\end{table*}

\subsection{Generalization on Other Vision-Language Models} 

In this paper, we mainly focus on the CLIP pre-trained model and use its open-source version with ViT-B/16 as the image encoder. Here we investigate whether CLIPood is general for other backbones or other variants of vision-language models. For other backbones, we use the CLIP pre-trained model with ResNet-50 as its image encoder (CLIP-RN50). For other variants of vision-language models, we use DeCLIP~\cite{cite:Arxiv2021DeCLIP} and Slip~\cite{cite:ECCV2022Slip}, and use their open-source pre-trained models with ViT-B/32 as the image encoders. We evaluate the performance of CLIPood on the OfficeHome dataset with both domain shift and open classes as the protocol in Section~\ref{sec:ODG}. As shown in Figure \ref{fig:backbone}, CLIPood consistently outperforms the zero-shot prediction on these three models, demonstrating its generalization ability on different architectures and variants of vision-language models.

\subsection{\cmr{Parameter Sensitivity}}
To investigate the robustness of hyper-parameters, we conduct detailed analysis experiments with different values of $\lambda$ and $\beta$. We experiment on the DomainNet dataset and the results are shown in Table \ref{table:sensitivity}. As shown, only if $\lambda$ is set too large to let the margin term dominate the loss will it hurt the performance. In a wide range, CLIPood on various values of hyper-parameters performs robustly and still outperforms the baseline results (59.9\% on DomainNet).

\begin{table}[ht]
    \centering
    \caption{Performance with different $\lambda$ and $\beta$ on DomainNet dataset.}
    \label{table:sensitivity}
    \begin{sc}
    \begin{subtable}[ht]{0.5\textwidth}
        \centering
        \begin{tabular}{lccccc}
        \toprule
        $\lambda$ & 0.01 & 0.03 & 0.1  & 0.3  & 1.0  \\
        \midrule
        CLIPood   & 62.9 & 63.1 & \textbf{63.3} & \textbf{63.5} & 61.8 \\
        \bottomrule
        \end{tabular}
    \end{subtable}
    
    \ \begin{subtable}[ht]{0.5\textwidth}
        \centering
        \begin{tabular}{lccccc}
        \toprule
        $\beta$ & 0.1  & 0.3  & 0.5  & 0.7  & 0.9  \\
        \midrule
        CLIPood   & 63.0 & \textbf{63.7} & \textbf{63.5} & \textbf{63.5} & 63.1 \\
        \bottomrule
        \end{tabular}
    \end{subtable}
    \end{sc}
\end{table}

\subsection{\cmr{Analysis on Zero-shot Performance}}
To further analyze how our approach exploits CLIP's pre-trained knowledge, we conduct an analysis on zero-shot performance. We compare the improvement of ERM and our approach on classes and domains where CLIP's zero-shot performance is poor. We first select the two domains where CLIP has the worst zero-shot performance as in Table \ref{table:worst2domain}. On the worst domain Quickdraw, our CLIPood still slightly outperforms ERM, while on the second worst domain Infograph, CLIPood outperforms ERM significantly.

\begin{table}[ht]
\centering
\caption{Performance on the worst 2 domains.}
\label{table:worst2domain}
\begin{sc}
\begin{tabular}{lcc}
\toprule
Method    & Quickdraw     & Infograph                      \\
\midrule
Zero-shot & 13.8          & 46.7                           \\
ERM       & 19.7          & 47.6                           \\
CLIPood   & \textbf{20.7} & \textbf{54.7} \\
\bottomrule
\end{tabular}
\end{sc}
\end{table}

Then, we investigate class-wise performance on the domain Infograph. Since there exist over 300 classes, we can only show part of them. According to zero-shot performance, we show 3 groups of classes with 12 classes in total. As shown in Table \ref{table:12class}, for the first group of classes where CLIP has poor zero-shot performance (class 1-4), CLIPood and ERM may lead to minor improvements upon zero-shot, or may outperform each other in different cases. This is because the pre-trained knowledge may not help certain classes. However, for the classes where the zero-shot performance is moderate (class 5-8), and where the zero-shot performance is near the average performance on the dataset (class 9-12), we find CLIPood, which benefits from better exploitation of pre-trained knowledge, generally provides better accuracies.

\begin{table}[ht]
\centering
\caption{Performance on 12 different classes.}
\label{table:12class}
\begin{sc}
\begin{tabular}{lcccccccccccc}
\toprule
\multirow{2}*{Infograph} & \multicolumn{4}{c}{low zero-shot acc} & \multicolumn{4}{c}{medium zero-shot acc} & \multicolumn{4}{c}{high zero-shot acc} \\
\cmidrule(lr){2-5} \cmidrule(lr){6-9} \cmidrule(lr) {10-13} & 1 & 2 & 3 & 4 & 5 & 6 & 7 & 8 & 9 & 10 & 11 & 12 \\
\midrule
Zero-shot & \textbf{0.0}    & 1.3    & 2.8    & 4.6    & 20.8   & \textbf{21.2}   & 22.5   & 23.1   & 46.2   & 46.3    & \textbf{46.8}    & 47.0    \\
ERM       & \textbf{0.0}    & 1.3    & \textbf{11.1}   & 22.7   & \textbf{37.5}   & 7.7    & 17.5   & 61.5   & 46.2   & 53.7    & 40.4    & 49.4    \\
CLIPood   & \textbf{0.0}    & \textbf{3.9}    & 2.8    & \textbf{27.3}   & \textbf{37.5}   & 19.2   & \textbf{30.0}   & \textbf{65.4}   & \textbf{61.5}   & \textbf{56.1}    & 44.7    & \textbf{53.0}    \\
\bottomrule
\end{tabular}
\end{sc}
\end{table}


\end{document}